\documentclass[10pt,twocolumn,letterpaper]{article}

\usepackage{iccv}              % To produce the CAMERA-READY version
\usepackage{xcolor}
\usepackage{enumitem}
\usepackage{algorithm}% http://ctan.org/pkg/algorithms
\usepackage{algorithmic}
\usepackage{placeins}
\usepackage{amsmath,amssymb}
\definecolor{iccvblue}{rgb}{0.21,0.49,0.74}
\usepackage[pagebackref,breaklinks,colorlinks,allcolors=iccvblue]{hyperref}
\usepackage{pifont}
\usepackage{amsmath}
\usepackage{multirow}
\usepackage{makecell}
\usepackage{wrapfig}
\usepackage{arydshln}
\usepackage{graphicx}
\usepackage{xcolor}

\def\paperID{7650} % *** Enter the Paper ID here
\def\confName{ICCV}
\def\confYear{2025}

\title{
\vspace{-2mm}
\includegraphics[width=0.7cm]{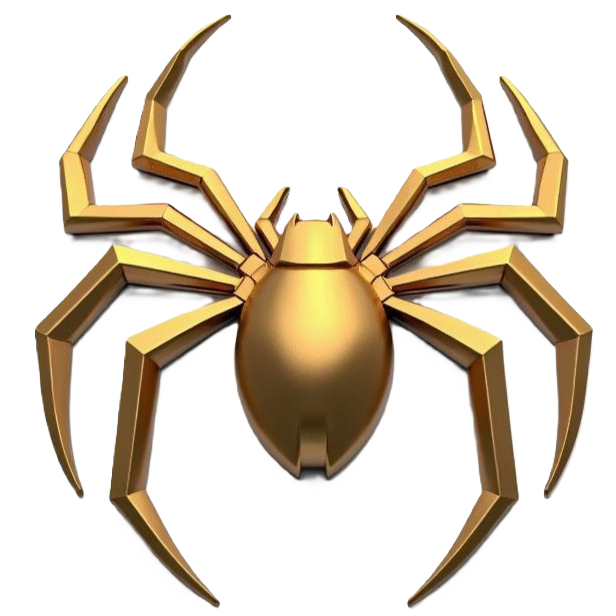}
Spider: Any-to-Many Multimodal LLM}

\author{Jinxiang Lai\\
HKUST\\
\and
Jie Zhang\\
HKUST\\
\and
Jun Liu\\
Tencent\\
\and
Jian Li\\
Tencent\\
\and
Xiaocheng Lu\\
HKUST\\
\and
Song Guo*\\
HKUST\\
}

\begin{document}
\maketitle
\begin{abstract}
Multimodal LLMs (MLLMs) have emerged as an extension of Large Language Models (LLMs), enabling the integration of various modalities. However, Any-to-Any MLLMs are limited to generating pairwise modalities 'Text + X' within a single response, such as Text + \{Image or Audio or Video\}.
To address this limitation, we introduce Spider, a novel efficient Any-to-Many Modalities Generation (AMMG) framework, which can generate an arbitrary combination of modalities 'Text + Xs', such as Text + \{Image and Audio and Video\}.
To achieve efficient AMMG, our Spider integrates three core components: a Base Model for basic X-to-X (i.e., Any-to-Any) modality processing, an Any-to-Many Instruction Template designed for producing Xs signal prompts, and a novel Efficient Decoders-Controller for controlling multimodal Decoders to generate Xs (many-modal) contents.
To train Spider, we constructed a novel Text-formatted Many-Modal (TMM) dataset, which facilitates learning the X-to-Xs (i.e., Any-to-Many) capability necessary for AMMG.
Ultimately, the well-trained Spider generates a pseudo X-to-Xs dataset, the first-ever X-to-Xs many-modal dataset, enhancing the potential for AMMG tasks in future research. 
Overall, this work not only pushes the boundary of multimodal interaction but also provides rich data support for advancing the field.
Code: \url{https://github.com/Layjins/Spider}
\end{abstract}    
\section{Introduction}
\label{sec:intro}
Large Language Models (LLMs) such as Vicuna \cite{vicuna}, LLaMA \cite{touvron2023llama}, ChatGPT \cite{chatgpt}, and GPT-4 \cite{gpt4} have demonstrated human-level proficiency in language understanding and generation. 
%These models, with their extensive training on large-scale corpora, have become foundational tools across a wide range of applications, from chatbots to advanced reasoning systems. 
However, as the demand for more complex, real-world applications grew, the need for integrating LLMs with multiple types of input and output modalities (e.g., text, images, audio, video) became apparent. This evolution has led to the rise of Multimodal LLMs (MLLMs), which extend LLMs' capabilities by incorporating multimodal perception modules \cite{huang2023language,zhu2023minigpt,su2022language,koh2023generating,alayrac2022flamingo,li2023blip,liu2023visual,su2023pandagpt,wu2023next}.

\begin{figure*}[t]
\vspace{-2mm}
\centering
\includegraphics[width=1.0\textwidth]{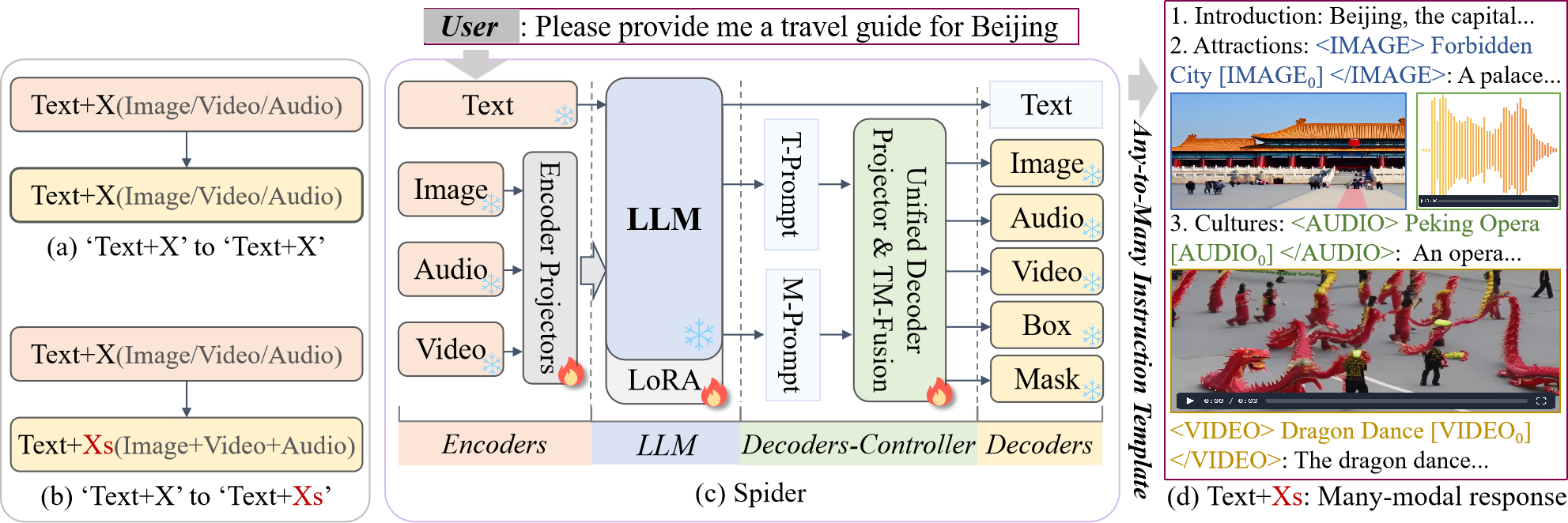}
\vspace{-4mm}
\caption{(a) The \textbf{X-to-X (Any-to-Any) MLLMs} support the input and output of pairwise modalities '\textbf{Text + X}'.
(b) Our \textbf{X-to-Xs (Any-to-Many) Spider model} produces many modalities '\textbf{Text + Xs}'. X denotes any-one-modality such as one of image or video or audio, and Xs means arbitrary-combination-modalities such as the combination of image and video and audio. A 'Text + Xs' example is shown in (d).
(c) The Spider structure comprises four parts including Encoders, LLM, Decoders-Controller, and Decoders.
The LLM is utilized as the core to process input multimodal information encoded by Encoders for semantic understanding and reasoning. 
Then, the LLM not only generates Text response, but also produces Text Prompt (T-Prompt) and Modality Prompt (M-Prompt) for the subsequent Decoders-Controller to control multimodal Decoders. 
(d) With Any-to-Many Instruction Template, T-Prompt and M-Prompt are gathered to form \textit{many-modal signal prompts}, which are able to control Decoders to generate many-modal contents. There is an example of many-modal signal prompts: '$<IMAGE>$ \textit{Forbidden City} $[{IMAGE}_0] </IMAGE>$. $<AUDIO>$ \textit{Peking Opera} $[{AUDIO}_0] </AUDIO>$.' Where '$<IMAGE> ... </IMAGE>$' and '$<AUDIO> ... </AUDIO>$' are the begin-end signal pairs of image and audio, respectively. '\textit{Forbidden City}' and '\textit{Peking Opera}' are T-Prompt. '$[{IMAGE}_0]$' and '$[{AUDIO}_0]$' are M-Prompt. Overall, we call this Modality-wise Grouping, i.e., each modality signal prompt is grouped by the corresponding begin-end signal pair containing the T-Prompt and M-Prompt inside. It allows arbitrary concatenation of different modality signal prompts.
}
\vspace{-2mm}
\label{fig:spider}
\end{figure*}

The development of MLLMs marks a significant advancement in enabling comprehensive understanding and generation across various modalities. Initially, models like LLaVA1.5 \cite{liu2023improved} and MiniGPT-4 \cite{zhu2023minigpt} were capable of processing only two modalities: text and images. 
%This setup, while useful for tasks such as image captioning and simple visual question answering, was limited in its ability to handle richer interactions involving more than two modalities. 
Further innovations saw the rise of models like PandaGPT \cite{su2023pandagpt}, OneLLM \cite{han2023onellm}, Gemini \cite{team2023gemini}, and NExT-GPT \cite{wu2023next}, which expanded support to four modalities, incorporating text, image, audio, and video into their multimodal frameworks. 
%Models like MiniGPT-v2 \cite{chen2023minigpt}, KOSMOS-2 \cite{peng2023kosmos}, and NExT-Chat \cite{zhang2023next} further enhanced capabilities by integrating region-level reasoning with modalities of box and mask.

However, as depicted in Fig.~\ref{fig:spider}(a), these \textbf{X-to-X (Any-to-Any) MLLMs} are restricted to generate pairwise modalities '\textbf{Text + X}' within a single interaction, such as 'Text + Image' or 'Text + Audio'. For example, when a user asks to generate an image of a dog, the model responds with image output. In subsequent interaction, to get an audio of the dog's bark, the user needs to give a new instruction. These MLLMs based on \textit{Multi-Round Dialogue Generation} paradigm, require several rounds of user instructions and do not allow for a seamless integration of multiple modalities within a single interaction. Each pair of modalities is handled independently, resulting in a fragmented user experience where the responses feel disjointed rather than cohesive.
Another example is in Fig.~\ref{fig:motivation_xs} of Appendix~\ref{sec:motivation_ammg}.

In contrast, as illustrated in Fig.~\ref{fig:spider}, our proposed \textbf{X-to-Xs (Any-to-Many) Spider model}, aims to achieve \textbf{Any-to-Many Modalities Generation (AMMG)} in a single response, which supports arbitrary combinations of a broader range of modalities: \textbf{text, image, audio, video, box, and mask}. For instance,  given a question ''Describe a dog using text, image, and audio.'' Our Spider can generate a cohesive output that combines text, images, and audio in a single response, greatly enhancing the user experience by providing comprehensive many-modal content all at once.
Another example is in Fig.~\ref{fig:spider}(d), generating a vivid travel guide.

The X-to-X MLLMs only require the LLM to perform X-modality instruction comprehension and prompt generation, which is a one-to-one task.
Differently, X-to-Xs is a more complex one-to-many task.
%Differently, X-to-Xs is a more complex one-to-many task, which needs to perform Xs-modality instruction comprehension and prompt generation.
Specifically, the X-to-Xs model needs the LLM to \textit{accurately understand the instructional requirements of any combination of Xs-modalities in the input question, and also produce the task prompts to correctly guide different decoders for Xs-modalities generation.}
For example, for the question “Generate an image of a dog, and I also would like to hear the dog’s bark,” due to the explicit appearance of “image,” the LLM may interpret it only as an image generation task, potentially overlooking the audio generation of “dog’s bark” or wrongly outputting the image generation task prompt for “dog’s bark”.
Fig.~\ref{fig:vis_compare} shows NExtGPT (X-to-X model) fails to follow user instruction to generate many-modal at a single response.

To address the above challenges and achieve \textit{efficient Any-to-Many Modalities Generation}, we designed a novel model named Spider as presented in Fig.~\ref{fig:spider}(c), and then constructed a novel Text-formatted Many-Modal (TMM) dataset to train this model. 
Our Spider incorporates three key components, i.e., Base Model, Any-to-Many Instruction Template, and Efficient Decoders-Controller: 

\noindent$\bullet$ \textbf{Base Model} (Encoders-LLM-Decoders structure), supports the basic X-to-X modality processing: Using multimodal Encoders to encode multimodal inputs, followed by an LLM to perform semantic understanding and reasoning, and finally the produced prompts by LLM are used to control the multimodal Decoders for generation. 

\noindent$\bullet$ \textbf{Any-to-Many Instruction Template}: \textit{To enable the LLM to understand multimodal instructions and produce many-modal signal prompts, thereby achieving accurate Any-to-Many Modalities Generation, we design an Any-to-Many Instruction Template applying the proposed Modality-wise Grouping rule.} An example of the proposed many-modal signal prompts is presented in Fig.~\ref{fig:spider}(d).

\noindent$\bullet$ \textbf{Efficient Decoders-Controller}, \textit{enables the LLM to effectively and efficiently control multiple task decoders for generating many-modal contents}:
(a) Obtain rich modality information for accurate decoding, by fusing the dominant Text Prompt (T-Prompt) and the auxiliary Modality Prompt (M-Prompt).
(b) Effective retain the input modality information in M-Prompt, by introducing M-Reconstruction loss.
(c) Effective feature alignment between LLM and Decoders, by the proposed MoE-based Unified Decoder Projector.
(d) Efficient learning: Finetuning LLM to generate specific T-Prompt is easy to achieve, i.e., learning efficient. Besides, M-Prompt is auxiliary information for T-Prompt. Its learning difficulty is much lower than treating it as the only controlling information like NExT-GPT.
(e) Efficient structure: As shown in Fig.~\ref{fig:motivation_decoder}, we design a Unified Decoder Projector, instead of multiple projectors like NExT-GPT.

Then, we constructed \textit{a novel Text-formatted Many-Modal (TMM) dataset to train the Spider model, enabling it to learn the X-to-Xs capability, i.e., to achieve Any-to-Many Modalities Generation.}
Existing datasets are mostly in the form of 'Text + X', which does not satisfy the X-to-Xs capability. NextGPT has constructed multimodal multi-round dialogue datasets, but each response in the dialogue is still in the form of 'Text + X'. 
Therefore, we need to construct a new TMM dataset to achieve the X-to-Xs capability. In the TMM dataset, the input is in the form of 'Text + X', while the output is in the form of \textbf{Text-formatted Xs (TXs)}, that is text, containing many-modal signal prompts (an example is presented in Fig.~\ref{fig:spider}(d)).
%There is a complete example of a Text-formatted Xs output: ·Here is the generated image of a dog: $<IMAGE>$ \textit{dog} $[{IMAGE}_0] </IMAGE>$, and the dog's bark: $<AUDIO>$ \textit{dog’s bark} $[{AUDIO}_0] </AUDIO>$'.
Eventually, the TMM dataset contains three types of datasets for different usage in training: \textbf{T-to-TXs} dataset for T-to-Xs capability finetuning, \textbf{X-to-TXs} dataset for X-to-Xs capability finetuning, and \textbf{T-to-TXs instruction} dataset for T-to-Xs instruction finetuning.

Finally, we use the Spider model well-trained on the TMM (X-to-TXs) dataset to generate a new pseudo X-to-Xs dataset. 
This is a first-ever X-to-Xs many-modal dataset for the Any-to-Many Modalities Generation task, providing rich data support for future research.
The output form of TMM dataset is TXs (i.e., text only) without diverse modalities, while the pseudo X-to-Xs dataset contains arbitrary combination modalities.
With TMM dataset, our Spider can perform X-to-Xs generation, due to no need to train the multimodal Decoders.
With pseudo X-to-Xs dataset, the multimodal Decoders can be end-to-end finetuning with LLM if needed in future work, due to having the ground truth modalities to supervise the Decoders.
More details of pseudo X-to-Xs dataset are in Appendix~\ref{sec:pseudo_dataset}.

In summary, our contributions are listed below: 
\begin{itemize}
    \item Beyond Any-to-Any Modality Generation, we introduce a novel Any-to-Many Modalities Generation paradigm that enables each response to contain ''Text + Xs''.

    \item We propose a novel efficient AMMG framework named Spider, which can generate arbitrary combinations and quantities of modalities. To achieve this, Spider integrates a Base Model, a novel Efficient Decoders-Controller, and a designed Any-to-Many Instruction Template.

    \item  We design a novel Any-to-Many Instruction Template, which enables the LLM to produce many-modal signal prompts, thereby achieving accurate AMMG.

    \item  We propose a novel Efficient Decoders-Controller that enables the LLM to effectively and efficiently control multiple task decoders to generate many-modal contents, improving the performance of AMMG task.

    \item  We construct a novel Text-formatted Many-Modal (TMM) dataset to trian Spider, enabling it to learn the X-to-Xs capability, i.e., to achieve AMMG. 

    \item  A new pseudo X-to-Xs dataset is generated by the well-trained Spider model, which is a first-ever X-to-Xs many-modal dataset, providing rich data support for future research on the AMMG task.
\end{itemize}

\section{Methodology}

\begin{figure*}[t]
\vspace{-4mm}
\centering
\includegraphics[width=0.9\textwidth]{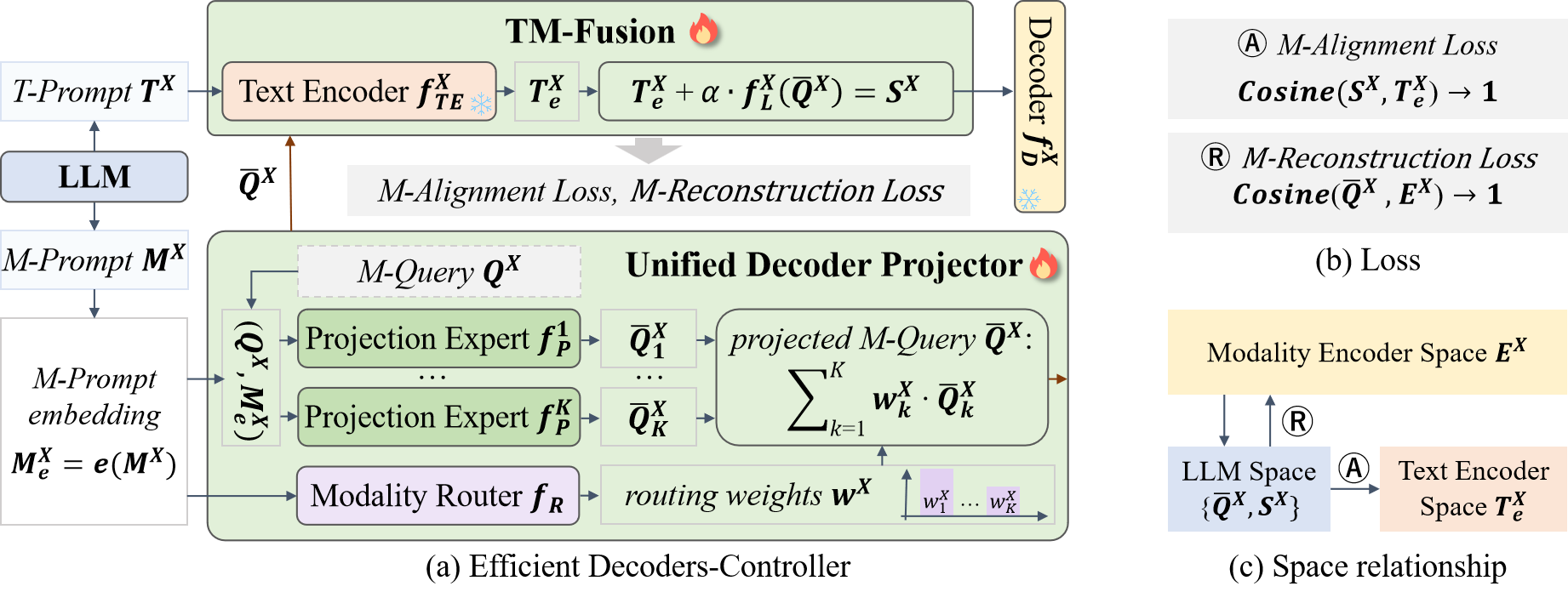}
\vspace{-3mm}
\caption{(a) Efficient Decoders-Controller consists of Unified Decoder Projector (UDP) and TM-Fusion (TMF), which enables the LLM to efficiently control multiple task Decoders to generate many-modal contents. $X$ means the variable or function corresponding to a specific X-modality, such as image or audio or video, etc. M-Prompt embedding $M_e^X=e(M^X)$ denotes obtaining the hidden embedding of $M^X$ extracted by LLM. (b) The M-Alignment Loss and M-Reconstruction Loss for optimizing the Decoders-Controller. (c) The intuitive embedding space relationship corresponding to (b).}
\vspace{-2mm}
\label{fig:decoder}
\end{figure*}

%\subsection{Problem Definition}
%We give a formal problem definition of Any-to-Many Modalities Generation.
%Let $\mathcal{X} = \{\text{Text}, X_1, X_2, \dots, X_n\}$ represent the set of available modalities, where each $X_i$ is a unique modality (e.g., image, audio, video, box, mask).
%For simplicity, we may alternatively use $X$ to represent any modality $X_i$.
%\textbf{Input Query}:  The input query $Q$ is defined as either a single text input or a combination of text with one additional modality: $Q = \{\text{Text}\} \quad \text{or} \quad Q = \{\text{Text}, X_i\}$.
%\textbf{Output Generation}: The output $Y$ is defined as a sequence containing any number of modalities from $\mathcal{X}$, structured as: $Y = (y_1, y_2, \dots, y_k), \, \text{where each } y_i \in \mathcal{X} \quad \text{and} \quad k \geq 1$. This allows $Y$ to be any arbitrary combination of modalities in response to $Q$.
%\textbf{Objective}: Given an input query $Q$, the objective is to generate a multimodal output $Y$ that accurately satisfies the instructional requirements of $Q$, integrating all requested modalities in a single response, avoiding multiple rounds of interaction to satisfy multimodal requirements.

\subsection{Model Architecture}
\label{sec:architecture}
As depicted in Fig.~\ref{fig:spider}(c), the Spider structure consists of four parts: Encoders, LLM, Decoders-Controller, and Decoders.
The LLM is utilized as the core to process input multimodal information encoded by Encoders for semantic understanding and reasoning. 
Then, the LLM not only generates Text response, but also produces Text Prompt (T-Prompt) and Modality Prompt (M-Prompt) for the subsequent Decoders-Controller to control multimodal Decoders. 

\noindent \textbf{Encoders.}
The inputs from diverse modalities are encoded by the pre-trained Encoders, where these encoded representations are projected into language-like representations that can be interpreted by the LLM. 
Here we adopt ImageBind \cite{girdhar2023imagebind} as the Encoders, which is a unified encoder supporting six different modalities. By leveraging ImageBind, we avoid the complexity of handling multiple heterogeneous encoders for various modalities. 
Then, using the Encoder Projectors which are linear projection layers, different input representations are aligned into the LLM space.

\noindent \textbf{LLM.}
In our Spider framework, we incorporate the open-source LLaMA2~\cite{touvron2023llama2} as the LLM component. 
The LLM receives representations from multiple modalities and performs semantic understanding and reasoning over these inputs. 
Beyond generating Text response, the LLM also produces Text Prompt (T-Prompt) and Modality Prompt (M-Prompt), which are utilized by the Decoders-Controller to control multimodal Decoders.

\noindent \textbf{Decoders-Controller.}
Decoders-Controller transforms the many-modal signal prompts (i.e., T-Prompt and M-Prompt) produced by LLM into representations that are understandable to subsequent multimodal Decoders.
The framework of Efficient Decoders-Controller is shown in Fig.~\ref{fig:decoder}, and it allows to effectively and efficiently control multiple Decoders to generate many-modal contents.

\noindent \textbf{Decoders.}
We leverage existing state-of-the-art latent-conditioned models for producing different modalities, specifically, Stable Diffusion v1.5 \cite{robin2022high} for image generation, AudioLDM \cite{LiuCYMLM0P23} for audio generation, Zeroscope v2 \cite{zeroscope} for video generation, Grounding DINO \cite{liu2023grounding} for bounding box prediction, and SAM \citep{kirillov2023sam} for mask prediction.

\subsection{Decoders-Controller}
As presented in Fig.~\ref{fig:decoder}, Efficient Decoders-Controller consists of Unified Decoder Projector (UDP) and TM-Fusion (TMF), which processes the many-modal signal prompts (i.e., T-Prompt and M-Prompt) produced by LLM, to efficiently control multiple task Decoders to generate many-modal contents.
We propose M-Alignment Loss and M-Reconstruction Loss to optimize the Decoders-Controller.

\begin{figure}[t]
%\vspace{-2mm}
\centering
\includegraphics[width=0.98\columnwidth]{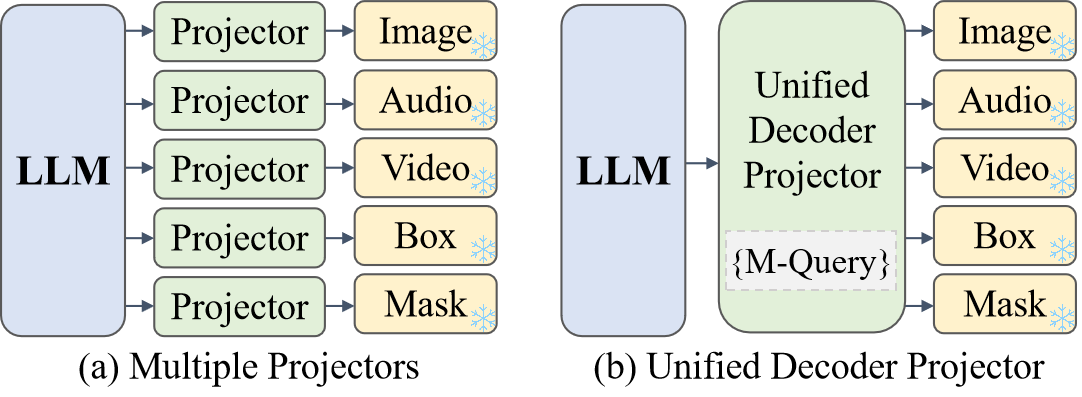}
\vspace{-3mm}
\caption{(a) Multiple Projectors (MP) for LLM-Decoders alignment. (b) Our Unified Decoder Projector (UDP).}
\vspace{-2mm}
\label{fig:motivation_decoder}
\end{figure}

\noindent \textbf{Unified Decoder Projector.} 
A Unified Decoder Projector (UDP) is designed to align the LLM with different Decoders.
\ding{172} As illustrated in Fig.~\ref{fig:decoder}(a), the UDP has $K$ \textit{Projection Experts} $\{f_P^k\}_{k=1, ..., K}$, where each projection expert is a stack of transformer layers. 
Empirically, in Fig.~\ref{fig:motivation_decoder}, $K=2$ is effective enough for our supported 5 modalities alignment with LLM, instead of using 5 multiple projectors in existing works, i.e., our UDP is more structure-efficient and scalable with the increasing of modalities.
\ding{173} To combine multiple projection experts within a single module, we introduce a dynamic \textit{Modality Router} $f_R$, designed to regulate the contribution of each expert and enhance the model’s capacity. The Modality Router $f_R$ is implemented as a multi-layer perception, which processes input embeddings and computes routing weights for each expert.
With the benefit of the MoE structure, UDP has a larger representation capacity leading to consistent performance improvements compared to MP in Fig.~\ref{fig:motivation_decoder}(a). 
\ding{174} Besides, we define learnable \textit{Modality Query} (M-Query) $\{Q^X\}_{X \in \mathcal{X}}$ for the corresponding output modalities, where $\mathcal{X}$ is the set of modalities, and $Q^X \in \mathbb{R}^{N^X\times D}$ contains $N^X$ tokens of dimension $D$. 
The parameter of M-Query is far less than a projection expert.
\ding{175} For modality $X$, the concatenation of M-Query $Q^X$ and M-Prompt embedding $M_e^X \in \mathbb{R}^{L^X \times D}$ are processed by UDP $f_{UDP}$, obtaining the projected M-Query $\bar Q^X \in \mathbb{R}^{N^X\times D}$:
\begin{align}
{\bar Q^X} &= f_{UDP}(Q^X, M_e^X) = \sum_{k=1}^K {w}^X_k \cdot {\bar Q^X_k}, \\
{w}^X &= \{{w}^X_k\}_{k=1,...,K} = \sigma [f_R(M_e^X)], \\
{\bar Q^X_k} &= f_P^k(Q^X, M_e^X),
\end{align}
where, ${w}^X \in \mathbb{R}^{K}$ is routing weights for $K$ experts, and $\sigma$ is softmax to ensure $\sum_{k=1}^K {w}^X_k=1$. 
After the projection by UDP, the M-Prompt embedding $M_e^X$ in LLM space, is transformed into the projected M-Query $\bar Q^X$ that is understandable to Decoder $f_D^X$.

\noindent \textbf{TM-Fusion.} 
As depicted in Fig.~\ref{fig:decoder}(a), the TM-Fusion (TMF) module integrates T-Prompt $T^X$ and the projected M-Query $\bar Q^X$ originally obtained from M-Prompt $M^X$, which aims to efficiently and accurately control Decoder $f_D^X$ for content generation.
Formally, TMF $f_{TMF}$ outputs the controlling embedding $S^X$ for Decoder:
\begin{align}
S^X &= f_{TMF}(T^X, {\bar Q^X}) = T^X_e + \alpha \cdot f^X_L({\bar Q^X}), \\
T^X_e &= f_{TE}^X(T^X),
\end{align}
where, $f_{TE}$ is the Text Encoder (i.e., the original conditional encoder in Decoder $f_D^X$), $f^X_L$ is a linear layer to align the embedding dimension of ${\bar Q^X}$ with $T^X_e$, and $\alpha$ is a constant variable to adjust the fusion weights and empirically set $\alpha$ = $0.2$.
The controlling embedding $S^X$ combines the information from T-Prompt and M-Prompt, by fusing $T^X_e$ and ${\bar Q^X}$.
Since LLM generating the required T-Prompt is easily achievable, $T^X_e$ encoded form T-Prompt by the Text Encoder can efficiently control the Decoder for content generation.
Besides, ${\bar Q^X}$ obtained from M-Prompt supplements the textual information of $T^X_e$ to retain input modality information, which is vital for realizing a more accurate modality generation.

\noindent \textbf{Loss of Decoders-Controller.} 
As shown in Fig.~\ref{fig:decoder}(b), we propose the M-Alignment Loss and M-Reconstruction Loss to optimize the Decoders-Controller.
\ding{172} The M-Alignment loss is expressed as $Cosine(S^X, T^X_e) \rightarrow 1$, which aims to maximize the cosine similarity between $S^X$ and $T^X_e$, i.e., let $S^X$ be semantic similar to $T^X_e$. It ensures $S^X$ is understandable by the Decoder.
\ding{173} The M-Reconstruction loss is expressed as $Cosine({\bar Q^X}, E^X) \rightarrow 1$, which aims to maximize the cosine similarity between ${\bar Q^X}$ and $E^X$, i.e., let ${\bar Q^X}$ be semantic similar to $E^X$. Where $E^X$ is the input-side modality embedding encoded by the modality Encoder (i.e., ImageBind). The M-Reconstruction loss is applied to not only retain the input modality information, but also prevent ${\bar Q^X}$ from collapsing toward zero in the M-Alignment loss.
\ding{174} The intuitive embedding space relationships are briefly illustrated in Fig.~\ref{fig:decoder}(c).

\begin{figure}[th]
\vspace{-2mm}
\centering
\includegraphics[width=1.0\columnwidth]{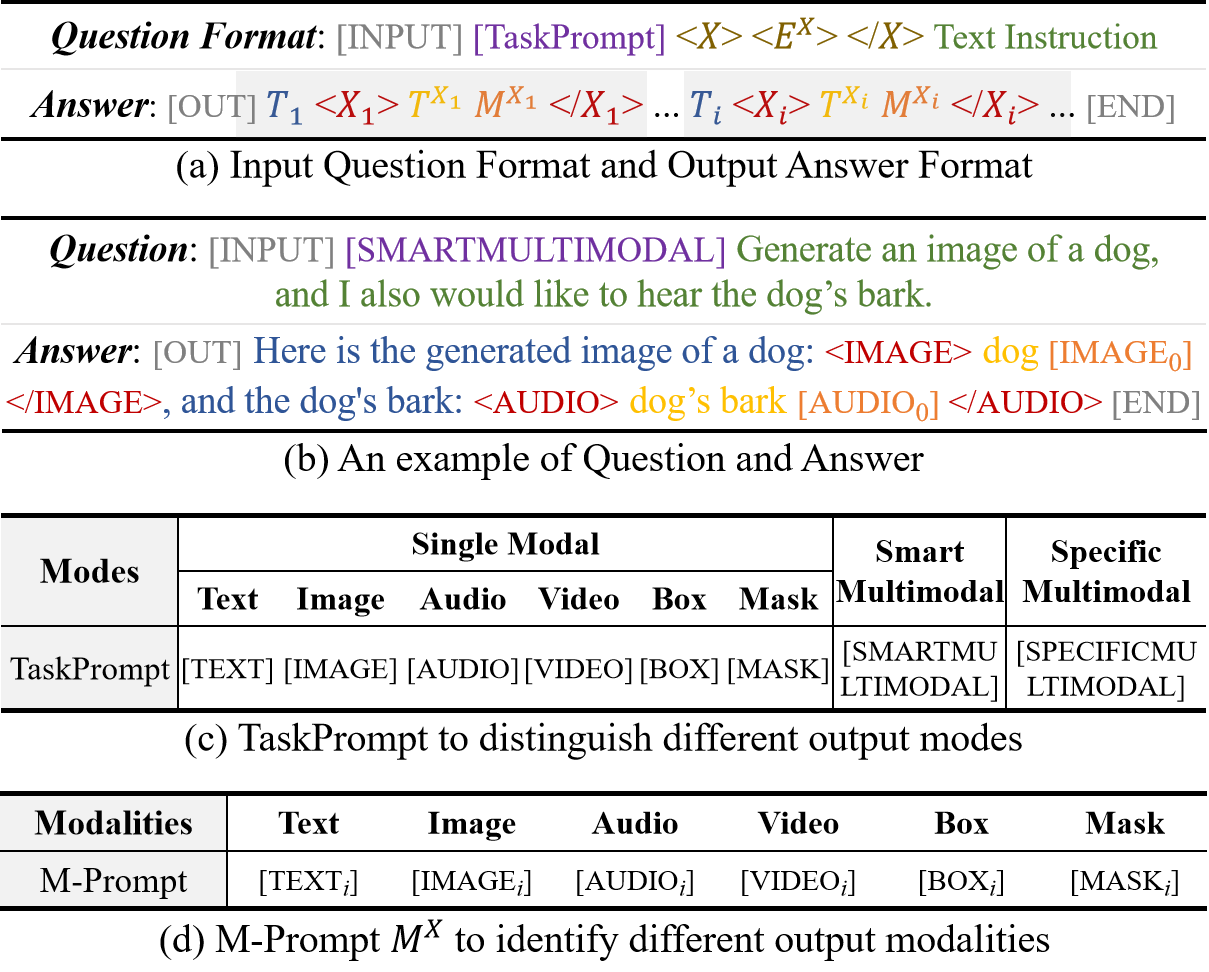}
\vspace{-6mm}
\caption{Any-to-Many Instruction Template. (a) Input Question Format and Output Answer Format. (b) An example of Question and Answer. Best view in color corresponding to (a). (c) TaskPrompt to distinguish different output modes. (d) M-Prompt $M^{X}$ to identify different output modalities, where $i$ is set to 0.}
\vspace{-2mm}
\label{fig:template}
\end{figure}

\subsection{Any-to-Many Instruction Template}
As shown in Fig.~\ref{fig:template}, we design a novel Any-to-Many Instruction Template, which enables the LLM to understand multimodal instructions and produce many-modal signal prompts, thereby achieving accurate Any-to-Many Modalities Generation.

\noindent \textbf{Input Question Format.} 
As presented in Fig.~\ref{fig:template}(a), the Input Question Format consists of four parts, including [INPUT], [TaskPrompt], $<X><E^X></X>$, and Text Instruction. An example is given in Fig.~\ref{fig:template}(b). \ding{172} [INPUT] is the question start signal. \ding{173} [TaskPrompt] specifies the output modes or tasks, making each task distinguishable. As shown in Fig.~\ref{fig:template}(c), we define three categories of output modes including Single Modal, Smart Multimodal, and Specific Multimodal, which contains a total of eight kinds of TaskPrompt. Single Modal mode means Spider only outputs one specific X-modality. Smart Multimodal mode can output any arbitrary combination of modalities. Specific Multimodal mode enables the user to construct an input question following the Answer Format, then our Spider model will output the corresponding multimodal answer.
\ding{174} $<X><E^X></X>$ indicates the input X-modality embedding $E^X$ is wrapped within the begin-end signal pairs $<X>...</X>$. \ding{174} Text Instruction is the text format user instruction.

\noindent \textbf{Output Answer Format.} 
As presented in Fig.~\ref{fig:template}(a), the Output Answer Format consists of three parts, including [OUT], $T_i <X_i> T^{X_i} \: M^{X_i} </X_i>$, and [END]. An example is given in Fig.~\ref{fig:template}(b).
\ding{172} [OUT] and [END] are the start and end signals of the answer, respectively.
\ding{173} $T_i <X_i> T^{X_i} \: M^{X_i} </X_i>$ forms a $X_i$ \textit{modality group} based on Modality-wise Grouping, where $T_i$ is the text response, $<X_i> ... </X_i>$ are the begin-end signal pairs, $T^{X_i}$ is T-Prompt, and $M^{X_i}$ is M-Prompt which serves as a modality signal to schedule the corresponding task decoder.
As shown in Fig.~\ref{fig:template}(c), each modality has a corresponding M-Prompt.
Based on the proposed Modality-wise Grouping, each $X_i$ modality group is gathered by the corresponding begin-end signal pair containing the T-Prompt and M-Prompt inside, and adjacency with the text response. This allows arbitrary concatenation of different modality groups.

\begin{figure}[t]
\vspace{-2mm}
\centering
\includegraphics[width=1.0\columnwidth]{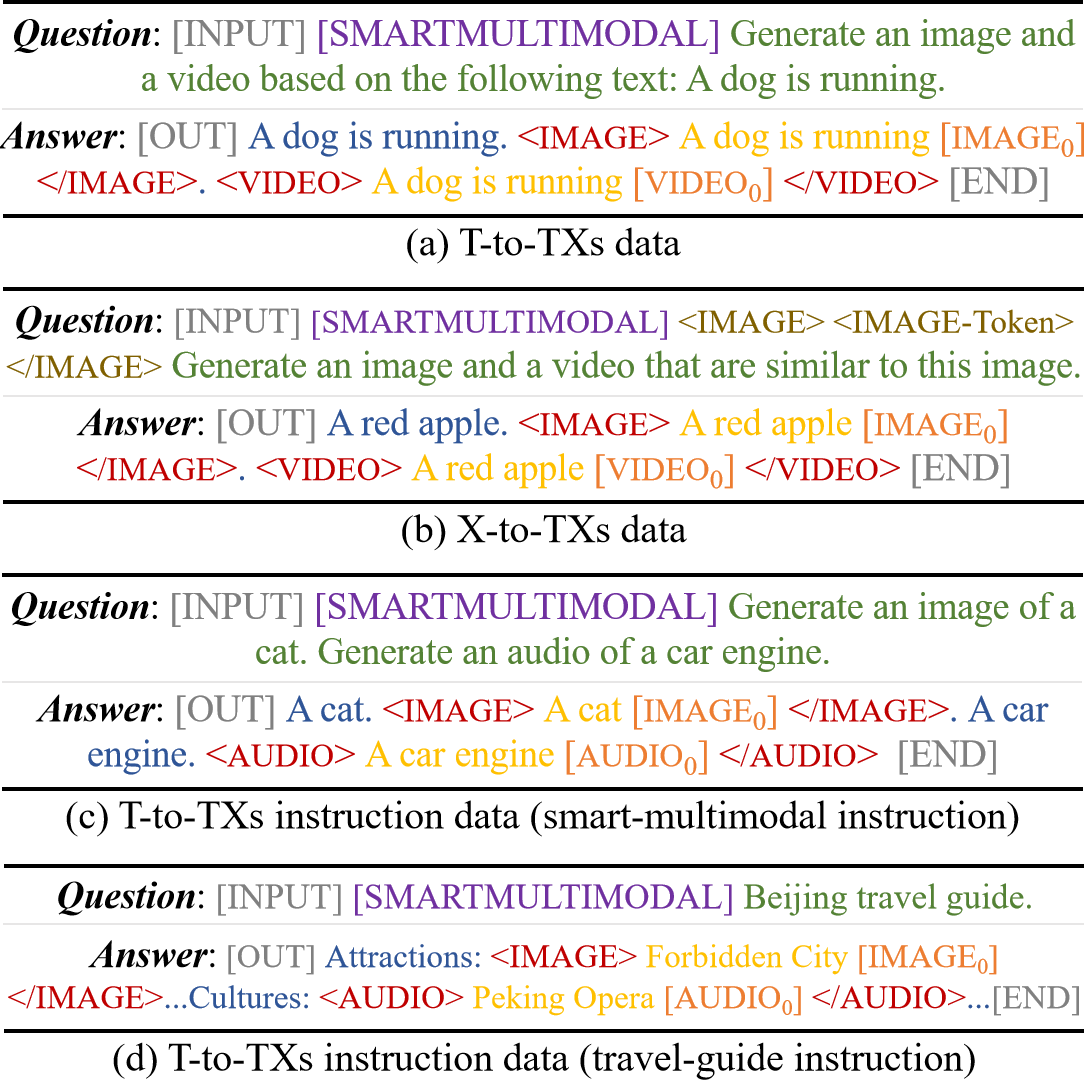}
\vspace{-6mm}
\caption{Examples of TMM dataset.}
\vspace{-2mm}
\label{fig:tmm}
\end{figure}

\begin{table*}[!ht]
\centering
\renewcommand{\tabcolsep}{2pt}
\renewcommand\arraystretch{1.1}
\small
\begin{tabular}{lccc|lcc|lcc}
\hline
\multicolumn{4}{c|}{\bf Image-to-Text on COCO-caption \cite{LinMBHPRDZ14}} & \multicolumn{3}{c|}{\bf Audio-to-Text on AudioCaps \cite{KimKLK19}} & \multicolumn{3}{c}{\bf Video-to-Text on MSR-VTT \cite{XuMYR16}}\\
\hline
\bf Method & \bf B@4($\uparrow$) & \bf METEOR($\uparrow$) & \bf CIDEr($\uparrow$) & \bf Method & \bf SPIDEr($\uparrow$) & \bf CIDEr($\uparrow$) & \bf Method & \bf B@4($\uparrow$) & \bf METEOR($\uparrow$)\\
\hline
OFA \cite{wang2022ofa} &  44.9 &  32.5 &  154.9 & CoDi \cite{abs-2305-11846} &  0.480 &  0.789 & mPLUG-2 \cite{XuYYSYXLBQWXZH023} &  57.8 &  34.9\\
\hdashline
NExT-GPT \cite{wu2023next} & 45.1  & 34.1  &  158.3 & NExT-GPT \cite{wu2023next} &  0.534  &  0.807  & NExT-GPT \cite{wu2023next} &  58.8 & 39.6 \\
\bf Our Spider & \bf 45.9   & \bf 34.3  & \bf 158.8  & \bf Our Spider &  \bf  0.537 & \bf  0.819 & \bf Our Spider & \bf 59.8 &  \bf 40.2\\
\hline
\end{tabular}
\vspace{-5pt}
\caption{Experimental comparisons on X-to-Text generation.
}
\vspace{-2pt}
\label{tab:exp_x2text}
\end{table*}

\begin{table*}[!ht]
\centering
\renewcommand{\tabcolsep}{7pt}
\renewcommand\arraystretch{1.1}
\small
\begin{tabular}{lc|lcc|lcc}
\hline
\multicolumn{2}{c|}{\bf Text-to-Image on COCO-caption \cite{LinMBHPRDZ14}} & \multicolumn{3}{c|}{\bf Text-to-Audio on AudioCaps \cite{KimKLK19}} & \multicolumn{3}{c}{\bf Text-to-Video on MSR-VTT \cite{XuMYR16}}\\
\hline
\bf Method & \bf FID ($\downarrow$) & \bf Method & \bf FD ($\downarrow$) & \bf IS ($\uparrow$) & \bf Method & \bf FID ($\downarrow$) & \bf CLIPSIM ($\uparrow$)\\
\hline
SD \cite{robin2022high} & 11.21 & CoDi \cite{abs-2305-11846} & \bf 22.90 & 8.77 & MakeVideo \cite{abs-2209-14792} & 	13.17 & 
 0.3049\\
\hdashline
NExT-GPT \cite{wu2023next} & 11.18 & NExT-GPT \cite{wu2023next} & 23.25 & 8.67 & NExT-GPT \cite{wu2023next} &  12.69 &  0.3197\\
\bf Our Spider & \bf 11.13 & \bf Our Spider & 23.02 & \bf 8.84 & \bf Our Spider & \bf 12.62 &  \bf 0.3258\\
\hline
\end{tabular}
\vspace{-5pt}
\caption{Experimental comparisons on Text-to-X generation.}
\vspace{-2pt}
\label{tab:exp_text2x}
\end{table*}

\begin{table*}[!ht]
\centering
\renewcommand{\tabcolsep}{7pt}
\renewcommand\arraystretch{1.1}
\small
\begin{tabular}{lcc|lc|lcc}
\hline
\multicolumn{3}{c|}{\bf Image-to-Image on COCO \cite{LinMBHPRDZ14}} & \multicolumn{2}{c|}{\bf Audio-to-Audio on VCTK \cite{veaux2017cstr}} & \multicolumn{3}{c}{\bf Video-to-Video on DAVIS \cite{PerazziPMGGS16}}\\
\hline
\bf Method & \bf CLIP ($\uparrow$) & \bf FID ($\downarrow$) & \bf Method & \bf MCD ($\downarrow$) & \bf Method & \bf CLIP-T ($\uparrow$) & \bf CLIP-I ($\uparrow$)\\
\hline
PFB-Diff \cite{abs-2306-16894} &\bf  30.81 &  5.93 & AudioLDM-L \cite{LiuCYMLM0P23}  & 0.349 & Pix2Video \cite{abs-2303-12688} & \bf 0.2891  & \bf 0.9767\\
\hdashline
NExT-GPT \cite{wu2023next} & 29.32   & 6.62 & NExT-GPT \cite{wu2023next} &  0.300 & NExT-GPT \cite{wu2023next} & 0.2684 &  0.9647\\
\bf Our Spider & 30.52   & \bf 5.33  & \bf Our Spider &  \bf 0.279 & \bf Our Spider & 0.2782 &  0.9715\\
\hline
\end{tabular}
\vspace{-5pt}
\caption{Experimental comparisons on X-to-X generation.
}
\vspace{-2pt}
\label{tab:exp_x2x}
\end{table*}

\subsection{TMM dataset}
We constructed a new Text-formatted Many-Modal (TMM) dataset to train the Spider model, enabling it to learn the X-to-Xs capability, i.e., to achieve Any-to-Many Modalities Generation. 
In the TMM dataset, the input is in the form of 'Text' or 'Text + X', which follows the Input Question Format. The output is in the form of \textit{Text-formatted Xs (TXs)}, that is text, containing many-modal signal prompts.
As illustrated in Fig.~\ref{fig:template} (a), the Output Answer Format is the TXs format.
Eventually, the TMM dataset contains three types of datasets for different usage in training: \textit{T-to-TXs} dataset for T-to-Xs capability finetuning, \textit{X-to-TXs} dataset for X-to-Xs capability finetuning, and \textit{T-to-TXs instruction} dataset for T-to-Xs instruction finetuning.
We show some examples in Fig.~\ref{fig:tmm}.
More details are in Appendix~\ref{sec:tmm_dataset}.

%\noindent \textbf{T-to-TXs and X-to-TXs Datasets.} 
%Based on the existing datasets including CC3M \cite{SoricutDSG18} (Image-Text), COCO \cite{LinMBHPRDZ14} (Image-Box and Image-Mask), AudioCap \cite{KimKLK19} (Audio-Text), and Webvid \cite{BainNVZ21} (Video-Text), we design corresponding Any-to-Many Instruction Templates for each task to construct the T-to-TXs dataset and X-to-TXs dataset.
%The T-to-TXs dataset contains three sub-datasets constructed from CC3M, AudioCap, and Webvid.
%The X-to-TXs dataset consists of five sub-datasets constructed from CC3M, COCO (Image-Box), COCO (Image-Mask), AudioCap, and Webvid.

%\noindent \textbf{T-to-TXs Instruction Dataset.} 
%Three T-to-TXs Instruction sub-datasets are constructed following the Any-to-Many Instruction Template format, including the smart-multimodal instruction sub-dataset based on Webvid, the specific-multimodal instruction sub-dataset based on Webvid, and the travel-guide instruction sub-dataset based on GPT-4o.
%The smart-multimodal and specific-multimodal instruction sub-datasets, concatenate multiple samples from Webvid, to mimic the arbitrary combination of output modalities.
%The travel-guide instruction sub-dataset is constructed with the assistance of GPT-4o, including \jinxiang{1000} travel guides for cities around the world.

\subsection{Spider Training}
The training process consists of three stages, including X-to-X Pretraining, X-to-TXs Finetuning, and X-to-TXs Instruction Finetuning.
\ding{172} The X-to-X Pretraining enables Spider to perform the basic X-to-X generation, connecting the four parts of Spider including Encoders, LLM, Decoders-Controller, and Decoders. 
%As shown in Fig.~\ref{fig:spider}, we only train the input-side Encoder Projectors, LoRA of LLM, and output-side Decoders-Controller, while other parameters are frozen.
%In this stage, we employ X-to-X tasks for training, including X-to-Text generation, Text-to-X generation, Image-to-Box prediction, and Image-to-Mask prediction.
\ding{173} The X-to-TXs Finetuning enables Spider to have the basic ability of X-to-Xs generation, via finetuning the LoRA of LLM with the proposed T-to-TXs and X-to-TXs Datasets.
\ding{174} The X-to-TXs Instruction Finetuning makes Spider achieve X-to-Xs generation in a proper manner, i.e., faithfully following and understanding the user instructions and generating desired many-modal outputs. 
%We further finetune the LoRA of LLM using the proposed T-to-TXs Instruction Dataset, T-to-TXs and X-to-TXs Datasets.
More details are in Appendix~\ref{sec:spider_train}.

\begin{table*}[!ht]
\centering
\renewcommand{\tabcolsep}{6pt}
\renewcommand\arraystretch{1.1}
\small
\begin{tabular}{l|ccc|ccc|ccc|cccc}
\hline
\multirow{2}*{\bf Method}  &  \multicolumn{3}{c|}{\bf $X_I$-to-Xs} & \multicolumn{3}{c|}{\bf $X_A$-to-Xs} & \multicolumn{3}{c|}{\bf $X_V$-to-Xs} &  \multicolumn{3}{c}{\bf Text-to-Xs} \\
\cline{2-14}
& I ($\downarrow$) & A ($\downarrow$) & V ($\downarrow$) & I ($\downarrow$) & A ($\downarrow$) & V ($\downarrow$) & I ($\downarrow$) & A ($\downarrow$) & V ($\downarrow$) & I ($\downarrow$) & A ($\downarrow$) & V ($\downarrow$) & B@4 ($\uparrow$)\\
\hline
NExT-GPT & 12.52 & 39.12 & 32.41 & 26.13 & 28.00 & 40.02 & 32.02 &  48.17 & 23.66 & 35.16 & 60.53 & 37.68 & 32.4\\
Spider-base & 9.24 & 22.75 & 18.95 & 15.60 & 16.82 & 22.17 & 17.88 & 24.01 &  13.22 & 20.73  & 38.27 & 23.11 & 40.5 \\
Spider & \bf 5.11 & \bf 18.44 & \bf 15.77 & \bf 12.36 & \bf 12.23 & \bf 18.93 & \bf 14.53 & \bf 20.55  & \bf 10.03  & \bf 17.48 & \bf 33.02 & \bf 20.02& \bf 40.8\\
\hline
\end{tabular}
\vspace{-5pt}
\caption{Experimental comparisons on Any-to-Many Modalities Generation (AMMG) task.
}
\vspace{-2pt}
\label{tab:exp_ammg}
\end{table*}

\begin{table*}[!ht]
\centering
\renewcommand{\tabcolsep}{7.5pt}
\renewcommand\arraystretch{1.1}
\small
\begin{tabular}{c|l|ccc|ccc|ccc}
\hline
\multirow{2}*{\bf Group}  & \multirow{2}*{\bf Method}  & \multicolumn{3}{c|}{\bf Text-to-X} & \multicolumn{3}{c|}{\bf X-to-X} & \multicolumn{3}{c}{\bf $X_I$-to-Xs}\\
\cline{3-11}
&& T2I ($\downarrow$) & T2A ($\downarrow$) & T2V ($\downarrow$) & I2I ($\uparrow$) & A2A ($\downarrow$) & V2V ($\uparrow$) & I ($\downarrow$) & A ($\downarrow$) & V ($\downarrow$)\\
\hline
\multirow{1}*{0} &NExT-GPT  & 11.18 & 23.25 & 12.69 & 29.32 & 0.300 & 0.2684 &  12.52 & 39.12 & 32.41 \\
\hdashline
\multirow{2}*{1} &Spider(MP)  & 11.17 & 23.22 & 12.64 & 29.42 & 0.298 & 0.2698  & 5.94  & 18.57  & 15.81\\
&Spider(UDP)  & \bf 11.13 & \bf 23.02 & \bf 12.62 & \bf 30.52 & \bf 0.279 & \bf 0.2782 & \bf 5.11 & \bf 18.44 & \bf 15.77\\
\hdashline
\multirow{3}*{2} &Spider(M-Prompt)  & 14.07 & 28.14 & 15.45 & 22.19 & 0.424 & 0.2103  &  8.13 & 21.42 & 17.82\\
&Spider(T-Prompt)  & \bf 11.11 & 23.04 & 12.63 & 29.04 & 0.322 & 0.2559  & 6.03 & 18.73 & 15.92\\
&Spider(TMF)  & 11.13 & \bf 23.02 & \bf 12.62 & \bf 30.52 & \bf 0.279 & \bf 0.2782 &\bf 5.11 & \bf 18.44& \bf 15.77\\
\hdashline
\multirow{3}*{3} &Spider(K=1) & 11.18 & 23.27 & 12.66 & 29.34 & 0.299 & 0.2690  & 6.00 & 18.63 & 15.94\\
&Spider(K=2) & \bf 11.13 & 23.02 & \bf 12.62 & \bf 30.52 & 0.279 & \bf 0.2782 &5.11 &18.44 & \bf 15.77\\
&Spider(K=3) & 11.13 & \bf 23.00 & 12.63 & 30.49 & \bf 0.276 & 0.2779  & \bf 5.03 & \bf 18.41 & 15.81\\
\hdashline
\multirow{2}*{4} &Spider(w/o MRL) & 11.14 & 23.07 & 12.62 & 29.89 & 0.289 & 0.2737 & 5.73 & 18.50 & 15.85\\
&Spider(w MRL) & \bf 11.13 & \bf 23.02 & \bf 12.62 & \bf 30.52 & \bf 0.279 & \bf 0.2782 & \bf 5.11 & \bf 18.44 & \bf 15.77\\
\hline
\end{tabular}
\vspace{-5pt}
\caption{Ablation study.
Notations of metrics: 
T2I ($\downarrow$) is FID for Text-to-Image on COCO-caption,
T2A ($\downarrow$) is FD for Text-to-Audio on AudioCaps,
T2V ($\downarrow$) is FID for Text-to-Video on MSR-VTT,
I2I ($\uparrow$) is CLIP for Image-to-Image on COCO,
A2A ($\downarrow$) is MCD for Audio-to-Audio on VCTK,
V2V ($\uparrow$) is CLIP-T for Video-to-Video on DAVIS.
$X_I$-to-Xs task on X-to-TXs (I2T) dataset, is consistent with Tab.~\ref{tab:exp_ammg}.
}
\vspace{-2pt}
\label{tab:ablation_com}
\end{table*}

\begin{figure*}[t]
\vspace{-2mm}
\centering
\includegraphics[width=1.0\textwidth]{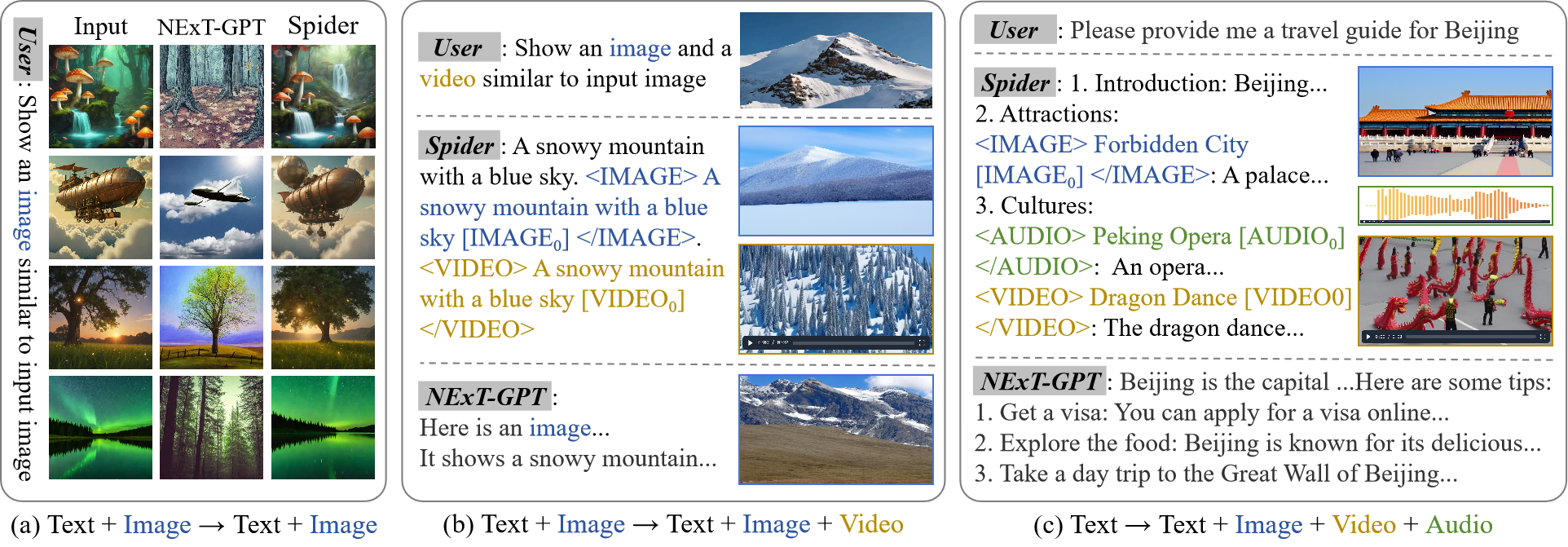}
\vspace{-6mm}
\caption{Qualitative examples.}
\vspace{-2mm}
\label{fig:vis_compare}
\end{figure*}

\section{Experiments}
\subsection{Any-to-Any Modality Generation}
Following \cite{wu2023next}, we evaluate our Spider on various benchmark tasks, including X-to-Text generation, Text-to-X generation, and Text-conditioned modality editing.
The results in Tab.~\ref{tab:exp_x2text}, Tab.~\ref{tab:exp_text2x}, and Tab.~\ref{tab:exp_x2x} show that our Spider is superior to NExT-GPT, meanwhile Spider obtains competitive performance compared to the state-of-the-art methods.
\textit{Comparisons on more task-specific datasets} are in Appendix~\ref{sec:task_spec}.

\noindent \textbf{X-to-Text Generation}
denotes the modality captioning tasks, of which the comparison results are shown in Tab.~\ref{tab:exp_x2text}.
Our Spider outperforms existing state-of-the-art methods, primarily due to it directly generates text through the LLM, leveraging the inherent expertise of the LLM.

\noindent \textbf{Text-to-X Generation}
denotes the text-conditioned modality synthesis tasks, of which the comparison results are shown in Tab.~\ref{tab:exp_text2x}.
Our Spider performs well compared to the state-of-the-art methods, and better than NExT-GPT.
The NExT-GPT heavily relies on the aligned modality embedding to control the text-conditioned Decoders, may resulting undesired generation.
Our Spider primarily relies on T-Prompt, with M-Prompt as an auxiliary aid, enabling more accurate control of the text-conditioned Decoders for modality generation.

\noindent \textbf{X-to-X Generation}
denotes the text-conditioned modality editing tasks, of which the comparison results are shown in Tab.~\ref{tab:exp_x2x}.
Our Spider is competitive compared to the state-of-the-art methods, and better than NExT-GPT.
Our Spider integrates T-Prompt and M-Prompt, where T-Prompt obtains the caption of the input modality, and M-Prompt can retain the input modality information due to the proposed Unified Decoder Projector and M-Reconstruction loss. Thus our Spider can achieve good performance on X-to-X tasks.

\subsection{Any-to-Many Modalities Generation}
\noindent \textbf{Datasets}. 
We use the constructed many-modal datasets for AMMG task evaluation, including X-to-TXs (I2T) dataset for $X_I$-to-Xs (i.e., Image-to-ManyModal) task, X-to-TXs (A2T) dataset for $X_A$-to-Xs (i.e., Audio-to-ManyModal) task, X-to-TXs (V2T) dataset for $X_V$-to-Xs (i.e., Video-to-ManyModal) task, and T-to-TXs TGI dataset for Text-to-Xs task.
The T-to-TXs TGI dataset consists of travel guides for cities, which is created with the assistance of GPT-4o.
More details are in Appendix~\ref{sec:tmm_dataset}.

\noindent \textbf{Metrics}. 
I ($\downarrow$) is FID metric for image generation, A ($\downarrow$) is FD for audio, V ($\downarrow$) is FID for video, and B@4 for text.
Each many-modal output is divided into different modality groups, and they are evaluated independently.
%Ground truth construction: Use the T-Prompt in TXs data as the text prompt, then generate modality data by text-control Decoders introduced in Sec.~\ref{sec:architecture} (Spider and NExT-GPT apply the same Decoders).

\noindent \textbf{Comparison Methods}. 
We compare Spider with Spider-base and NExT-GPT. NExT-GPT is an any-to-any generation model. For better comparison, we introduce Spider-base, an any-to-many generation model, which applies Multiple Projectors (MP) with M-Prompt only like NExT-GPT, instead of the Efficient Decoders-Controller that Spider did.
In other words, Spider-base is similar to NExT-GPT integrating our Any-to-Many Instruction Template.

\noindent \textbf{Performance}. 
As illustrated in Fig.~\ref{tab:exp_ammg}, the results show that:
(1) Our Spier and Spier-base outperform NExT-GPT on AMMG task by a large margin, because of Spider's ability to generate many-modal content Xs through the usage of the proposed Any-to-Many Instruction Template, while NExT-GPT can only produce single-modal X.
(2) Spider obtains consistent improvements compared to Spider-base, due to the benefits of TM-Fusion (TMF) fusing T-Prompt and M-Prompt obtaining rich modality information for accurate decoding, and MoE-based Unified Decoder Projector (UDP) achieving effective feature alignment between LLM and Decoders.
(3) Specifically, Spider has better performances compared to Spider-base on Text-to-Xs task on T-to-TXs TGI dataset, and on $X_i$-to-$X_j$ task (input and output modalities are different, e.g., Image-to-Audio), since T-Prompt dominates these tasks, while Spider-base only applied M-Prompt that can not be perfectly aligned.
On $X_i$-to-$X_i$ task (input and output modalities are the same, e.g., Image-to-Image), Spider obtains a good performance improvement, due to using T-Prompt and the help of modality info preserved in M-Prompt with effective feature alignment by UDP.

\section{Ablation Study}
\noindent \textbf{Influence of UDP.}
Tab.~\ref{tab:ablation_com} Group-1 shows that Spider(UDP) obtains consistent improvements compared to Spider(MP), due to UDP achieving more effective feature alignment between LLM and Decoders, which benefited from the MoE structure with larger representation capacity.
Specifically, Spider(UDP) obtains a decent improvement on X-to-X task and $X_i$-to-$X_i$ task on X-to-TXs (I2T) dataset, due to UDP achieving better feature alignment that helps to preserve more modality info in M-Prompt.
Spider(UDP) is slightly better than Spider(MP) on Text-to-X and $X_i$-to-$X_j$ tasks, since the influence of M-Prompt decreases on these tasks. 

\noindent \textbf{Influence of TMF.}
Tab.~\ref{tab:ablation_com} Group-2 indicates that :
(1) Spider(TMF) obtains great improvements by fusing the dominant T-Prompt and auxiliary M-Prompt, which obtains rich modality information for accurate decoding. 
(2) Similar to NExT-GPT, Spider(M-Prompt) only employed M-Prompt for generation. While the modality token numbers of Spider are (1, 1, 1) for (image, audio, video) compared to (4, 8, 24) of NExT-GPT, and the training iterations of Spider(M-Prompt) are 4 times less than NExT-GPT, leading to suboptimal performance. But with the help of T-Prompt, Spider(TMF) still obtains outstanding performance.
(3) On X-to-X task and $X_i$-to-$X_i$ task on X-to-TXs (I2T) dataset, Spider(TMF) obtains a good performance improvement, with the help of preserved modality info in M-Prompt.
Spider(TMF) has similar performances with Spider(T-Prompt) on Text-to-X task and $X_i$-to-$X_j$ task, since M-Prompt has a weak influence on these tasks. 

\noindent \textbf{Influence of Experts.}
Tab.~\ref{tab:ablation_com} Group-3 shows that the number of experts $K=2$ is enough to allow our Unified Decoder Projector to effectively align the LLM with the integrated 5 task Decoders. 
$K=2$ obtains good performance improvements than $K=1$ on X-to-X task and $X_i$-to-$X_i$ task on X-to-TXs (I2T) dataset, while $K=3$ shows minor performance gains than $K=2$. 
Since M-Prompt has a weak influence on Text-to-X task and $X_i$-to-$X_j$ task, the performances of different $K$ are similar.

\noindent \textbf{Influence of M-Reconstruction Loss (MRL).}
Tab.~\ref{tab:ablation_com} Group-4 shows that employing MRL achieves performance improvements on X-to-X task and $X_i$-to-$X_i$ task on X-to-TXs (I2T) dataset, because MRL is applied to not only retain the input modality information, but also prevent ${\bar Q^X}$ from collapsing toward zero in the M-Alignment loss.
On Text-to-X and $X_i$-to-$X_j$ tasks, the performances of Spider with and without MRL are similar, due to T-Prompt playing a dominant role in controlling decoders.

\section{Qualitative Analysis}
To demonstrate Spider's remarkable ability to generate arbitrary combinations of modalities within a single response, we provide qualitative comparisons in Fig.~\ref{fig:vis_compare}.
More examples are in Appendix~\ref{sce:vis} and our project page \url{https://anonymous.4open.science/r/spider}.
Fig.~\ref{fig:vis_compare}(a) shows that Spider generated the image that is more similar to the input image, compared to NExT-GPT. Because our Spider can obtain rich modality information for accurate decoding by fusing T-Prompt and M-Prompt, and can effectively preserve the visual details from the input image with the help of M-Reconstruction loss.
Fig.~\ref{fig:vis_compare} (b) and (c) demonstrate that Spider can accurately generate the many-modal content according to the user prompt, due to the applied Any-to-Many Instruction Template and Efficient Decoders-Controller. But NExT-GPT only generated one type of modality at a single response, i.e., failed to produce the many-modal content.
\section{Conclusion}
This paper presents significant advancements in multimodal generation through the introduction of any-to-many modalities generation paradigm, moving beyond traditional any-to-any modality generation. 
Our novel efficient any-to-many modalities generation framework, named Spider, allows for the seamless integration of diverse modality combinations within a single response.
Spider's key components are the proposed efficient Decoders-Controller and the designed Any-to-Many Instruction Template.
The Decoders-Controller enables the LLM to efficiently control multiple task decoders for generating many-modal contents.
The Any-to-Many Instruction Template enables the LLM to understand multimodal instructions and produce many-modal signal prompts, thereby achieving accurate any-to-many modalities generation.
Furthermore, a Text-formatted Many-Modal dataset is constructed, empowering Spider to learn X-to-Xs capability. We also generate a pseudo X-to-Xs dataset that provides valuable data support for future advancements in any-to-many modalities generation. 

{
    \small
    \bibliographystyle{ieeenat_fullname}
    \bibliography{main}

\begin{thebibliography}{73}
\providecommand{\natexlab}[1]{#1}
\providecommand{\url}[1]{\texttt{#1}}
\expandafter\ifx\csname urlstyle\endcsname\relax
  \providecommand{\doi}[1]{doi: #1}\else
  \providecommand{\doi}{doi: \begingroup \urlstyle{rm}\Url}\fi

\bibitem[gpt(2023)]{gpt4}
Gpt-4 technical report.
\newblock In \emph{OpenAI}, 2023.

\bibitem[Alayrac et~al.(2022)Alayrac, Donahue, Luc, Miech, Barr, Hasson, Lenc, Mensch, Millican, Reynolds, et~al.]{alayrac2022flamingo}
Jean-Baptiste Alayrac, Jeff Donahue, Pauline Luc, Antoine Miech, Iain Barr, Yana Hasson, Karel Lenc, Arthur Mensch, Katherine Millican, Malcolm Reynolds, et~al.
\newblock Flamingo: a visual language model for few-shot learning.
\newblock In \emph{NeurIPS}, 2022.

\bibitem[An et~al.(2023)An, Zhang, Yang, Gupta, Huang, Luo, and Yin]{Jie2023Latent}
Jie An, Songyang Zhang, Harry Yang, Sonal Gupta, Jia{-}Bin Huang, Jiebo Luo, and Xi Yin.
\newblock Latent-shift: Latent diffusion with temporal shift for efficient text-to-video generation.
\newblock In \emph{arXiv}, 2023.

\bibitem[Avrahami et~al.(2023)Avrahami, Fried, and Lischinski]{AvrahamiFL23}
Omri Avrahami, Ohad Fried, and Dani Lischinski.
\newblock Blended latent diffusion.
\newblock \emph{{ACM} Trans. Graph.}, 42\penalty0 (4):\penalty0 149:1--149:11, 2023.

\bibitem[Bain et~al.(2021)Bain, Nagrani, Varol, and Zisserman]{BainNVZ21}
Max Bain, Arsha Nagrani, G{\"{u}}l Varol, and Andrew Zisserman.
\newblock Frozen in time: {A} joint video and image encoder for end-to-end retrieval.
\newblock In \emph{Proceedings of the ICCV}, pages 1708--1718, 2021.

\bibitem[Brown et~al.(2020)Brown, Mann, Ryder, Subbiah, Kaplan, Dhariwal, Neelakantan, Shyam, Sastry, Askell, et~al.]{brown2020language}
Tom Brown, Benjamin Mann, Nick Ryder, Melanie Subbiah, Jared~D Kaplan, Prafulla Dhariwal, Arvind Neelakantan, Pranav Shyam, Girish Sastry, Amanda Askell, et~al.
\newblock Language models are few-shot learners.
\newblock In \emph{NeurIPS}, 2020.

\bibitem[Cerspense(2023)]{zeroscope}
Cerspense.
\newblock Zeroscope: Diffusion-based text-to-video synthesis.
\newblock 2023.

\bibitem[Ceylan et~al.(2023)Ceylan, Huang, and Mitra]{abs-2303-12688}
Duygu Ceylan, Chun{-}Hao~Paul Huang, and Niloy~J. Mitra.
\newblock Pix2video: Video editing using image diffusion.
\newblock In \emph{ICCV}, 2023.

\bibitem[Chiang et~al.(2023)Chiang, Li, Lin, Sheng, Wu, Zhang, Zheng, Zhuang, Zhuang, Gonzalez, Stoica, and Xing]{vicuna}
Wei-Lin Chiang, Zhuohan Li, Zi Lin, Ying Sheng, Zhanghao Wu, Hao Zhang, Lianmin Zheng, Siyuan Zhuang, Yonghao Zhuang, Joseph~E. Gonzalez, Ion Stoica, and Eric~P. Xing.
\newblock Vicuna: An open-source chatbot impressing gpt-4 with 90%* chatgpt quality.
\newblock 2023.

\bibitem[Chowdhery et~al.(2022)Chowdhery, Narang, Devlin, Bosma, Mishra, Roberts, Barham, Chung, Sutton, Gehrmann, et~al.]{chowdhery2022palm}
Aakanksha Chowdhery, Sharan Narang, Jacob Devlin, Maarten Bosma, Gaurav Mishra, Adam Roberts, Paul Barham, Hyung~Won Chung, Charles Sutton, Sebastian Gehrmann, et~al.
\newblock Palm: Scaling language modeling with pathways.
\newblock In \emph{arXiv}, 2022.

\bibitem[Couairon et~al.(2023)Couairon, Verbeek, Schwenk, and Cord]{CouaironVSC23}
Guillaume Couairon, Jakob Verbeek, Holger Schwenk, and Matthieu Cord.
\newblock Diffedit: Diffusion-based semantic image editing with mask guidance.
\newblock In \emph{Proceedings of the ICLR}, 2023.

\bibitem[Devlin et~al.(2019)Devlin, Chang, Lee, and Toutanova]{Devlin2019BERT}
Jacob Devlin, Ming-Wei Chang, Kenton Lee, and Kristina Toutanova.
\newblock Bert: pre-training of deep bidirectional transformers for language understanding.
\newblock In \emph{NAACL-HLT}, 2019.

\bibitem[Ding et~al.(2021)Ding, Yang, Hong, Zheng, Zhou, Yin, Lin, Zou, Shao, Yang, and Tang]{DingYHZZYLZSYT21}
Ming Ding, Zhuoyi Yang, Wenyi Hong, Wendi Zheng, Chang Zhou, Da Yin, Junyang Lin, Xu Zou, Zhou Shao, Hongxia Yang, and Jie Tang.
\newblock Cogview: Mastering text-to-image generation via transformers.
\newblock In \emph{Proceedings of the NeurIPS}, pages 19822--19835, 2021.

\bibitem[Feng et~al.(2022)Feng, He, Fu, Jampani, Akula, Narayana, Basu, Wang, and Wang]{feng2022training}
Weixi Feng, Xuehai He, Tsu-Jui Fu, Varun Jampani, Arjun Akula, Pradyumna Narayana, Sugato Basu, Xin~Eric Wang, and William~Yang Wang.
\newblock Training-free structured diffusion guidance for compositional text-to-image synthesis.
\newblock \emph{arXiv preprint arXiv:2212.05032}, 2022.

\bibitem[Ge et~al.(2023)Ge, Nah, Liu, Poon, Tao, Catanzaro, Jacobs, Huang, Liu, and Balaji]{ge2023preserve}
Songwei Ge, Seungjun Nah, Guilin Liu, Tyler Poon, Andrew Tao, Bryan Catanzaro, David Jacobs, Jia-Bin Huang, Ming-Yu Liu, and Yogesh Balaji.
\newblock Preserve your own correlation: A noise prior for video diffusion models.
\newblock In \emph{Proceedings of the IEEE/CVF International Conference on Computer Vision}, pages 22930--22941, 2023.

\bibitem[Gemmeke et~al.(2017)Gemmeke, Ellis, Freedman, Jansen, Lawrence, Moore, Plakal, and Ritter]{gemmeke2017audio}
Jort~F Gemmeke, Daniel~PW Ellis, Dylan Freedman, Aren Jansen, Wade Lawrence, R~Channing Moore, Manoj Plakal, and Marvin Ritter.
\newblock {AudioSet}: An ontology and human-labeled dataset for audio events.
\newblock In \emph{IEEE International Conference on Acoustics, Speech and Signal Processing}, pages 776--780. IEEE, 2017.

\bibitem[Girdhar et~al.(2023)Girdhar, El-Nouby, Liu, Singh, Alwala, Joulin, and Misra]{girdhar2023imagebind}
Rohit Girdhar, Alaaeldin El-Nouby, Zhuang Liu, Mannat Singh, Kalyan~Vasudev Alwala, Armand Joulin, and Ishan Misra.
\newblock Imagebind: One embedding space to bind them all.
\newblock In \emph{CVPR}, 2023.

\bibitem[Gontier et~al.(2021)Gontier, Serizel, and Cerisara]{GontierSC21}
F{\'{e}}lix Gontier, Romain Serizel, and Christophe Cerisara.
\newblock Automated audio captioning by fine-tuning {BART} with audioset tags.
\newblock In \emph{Proceedings of the DCASE}, pages 170--174, 2021.

\bibitem[Han et~al.(2024)Han, Gong, Zhang, Wang, Zhang, Lin, Qiao, Gao, and Yue]{han2023onellm}
Jiaming Han, Kaixiong Gong, Yiyuan Zhang, Jiaqi Wang, Kaipeng Zhang, Dahua Lin, Yu Qiao, Peng Gao, and Xiangyu Yue.
\newblock Onellm: One framework to align all modalities with language.
\newblock In \emph{CVPR}, 2024.

\bibitem[Hong et~al.(2022)Hong, Ding, Zheng, Liu, and Tang]{abs-2205-15868}
Wenyi Hong, Ming Ding, Wendi Zheng, Xinghan Liu, and Jie Tang.
\newblock Cogvideo: Large-scale pretraining for text-to-video generation via transformers.
\newblock In \emph{arXiv}, 2022.

\bibitem[Huang et~al.(2023{\natexlab{a}})Huang, Huang, Yang, Ren, Liu, Li, Ye, Liu, Yin, and Zhao]{HuangHY0LLYLYZ23}
Rongjie Huang, Jiawei Huang, Dongchao Yang, Yi Ren, Luping Liu, Mingze Li, Zhenhui Ye, Jinglin Liu, Xiang Yin, and Zhou Zhao.
\newblock Make-an-audio: Text-to-audio generation with prompt-enhanced diffusion models.
\newblock In \emph{ICML}, 2023{\natexlab{a}}.

\bibitem[Huang et~al.(2023{\natexlab{b}})Huang, Li, Yang, Shi, Chang, Ye, Wu, Hong, Huang, Liu, et~al.]{huang2023audiogpt}
Rongjie Huang, Mingze Li, Dongchao Yang, Jiatong Shi, Xuankai Chang, Zhenhui Ye, Yuning Wu, Zhiqing Hong, Jiawei Huang, Jinglin Liu, et~al.
\newblock Audiogpt: Understanding and generating speech, music, sound, and talking head.
\newblock In \emph{arXiv}, 2023{\natexlab{b}}.

\bibitem[Huang et~al.(2023{\natexlab{c}})Huang, Dong, Wang, Hao, Singhal, Ma, Lv, Cui, Mohammed, Liu, et~al.]{huang2023language}
Shaohan Huang, Li Dong, Wenhui Wang, Yaru Hao, Saksham Singhal, Shuming Ma, Tengchao Lv, Lei Cui, Owais~Khan Mohammed, Qiang Liu, et~al.
\newblock Language is not all you need: Aligning perception with language models.
\newblock In \emph{arXiv}, 2023{\natexlab{c}}.

\bibitem[Huang et~al.(2025)Huang, Tu, and Xu]{abs-2306-16894}
Wenjing Huang, Shikui Tu, and Lei Xu.
\newblock Pfb-diff: Progressive feature blending diffusion for text-driven image editing.
\newblock \emph{Neural Networks}, 2025.

\bibitem[Kim et~al.(2019)Kim, Kim, Lee, and Kim]{KimKLK19}
Chris~Dongjoo Kim, Byeongchang Kim, Hyunmin Lee, and Gunhee Kim.
\newblock Audiocaps: Generating captions for audios in the wild.
\newblock In \emph{Proceedings of the NAACL}, pages 119--132, 2019.

\bibitem[Kim et~al.(2022)Kim, Kim, Oh, Kim, Park, Sim, Lee, and Lee]{abs-2210-17143}
Eungbeom Kim, Jinhee Kim, Yoori Oh, Kyungsu Kim, Minju Park, Jaeheon Sim, Jinwoo Lee, and Kyogu Lee.
\newblock Improving audio-language learning with mixgen and multi-level test-time augmentation.
\newblock In \emph{arXiv}, 2022.

\bibitem[Kirillov et~al.(2023)Kirillov, Mintun, Ravi, Mao, Rolland, Gustafson, Xiao, Whitehead, Berg, Lo, Doll{\'a}r, and Girshick]{kirillov2023sam}
Alexander Kirillov, Eric Mintun, Nikhila Ravi, Hanzi Mao, Chloe Rolland, Laura Gustafson, Tete Xiao, Spencer Whitehead, Alexander~C. Berg, Wan-Yen Lo, Piotr Doll{\'a}r, and Ross Girshick.
\newblock Segment anything.
\newblock In \emph{arXiv}, 2023.

\bibitem[Koh et~al.(2023)Koh, Fried, and Salakhutdinov]{koh2023generating}
Jing~Yu Koh, Daniel Fried, and Ruslan Salakhutdinov.
\newblock Generating images with multimodal language models.
\newblock In \emph{arXiv}, 2023.

\bibitem[Krojer et~al.(2023)Krojer, Poole-Dayan, Voleti, Pal, and Reddy]{krojer2023diffusion}
Benno Krojer, Elinor Poole-Dayan, Vikram Voleti, Christopher Pal, and Siva Reddy.
\newblock Are diffusion models vision-and-language reasoners?
\newblock In \emph{Thirty-seventh Conference on Neural Information Processing Systems}, 2023.

\bibitem[Li et~al.(2023{\natexlab{a}})Li, Li, Savarese, and Hoi]{li2023blip}
Junnan Li, Dongxu Li, Silvio Savarese, and Steven Hoi.
\newblock Blip-2: Bootstrapping language-image pre-training with frozen image encoders and large language models.
\newblock In \emph{ICML}, 2023{\natexlab{a}}.

\bibitem[Li et~al.(2023{\natexlab{b}})Li, He, Wang, Li, Wang, Luo, Wang, Wang, and Qiao]{li2023videochat}
KunChang Li, Yinan He, Yi Wang, Yizhuo Li, Wenhai Wang, Ping Luo, Yali Wang, Limin Wang, and Yu Qiao.
\newblock Videochat: Chat-centric video understanding.
\newblock In \emph{arXiv}, 2023{\natexlab{b}}.

\bibitem[Li et~al.(2020)Li, Yin, Li, Zhang, Hu, Zhang, Wang, Hu, Dong, Wei, Choi, and Gao]{Li0LZHZWH0WCG20}
Xiujun Li, Xi Yin, Chunyuan Li, Pengchuan Zhang, Xiaowei Hu, Lei Zhang, Lijuan Wang, Houdong Hu, Li Dong, Furu Wei, Yejin Choi, and Jianfeng Gao.
\newblock Oscar: Object-semantics aligned pre-training for vision-language tasks.
\newblock In \emph{Proceedings of the ECCV}, pages 121--137, 2020.

\bibitem[Li et~al.(2023{\natexlab{c}})Li, Chu, Wu, Yuan, Liu, Zhang, Li, Feng, Ding, and Wang]{li2023videogen}
Xin Li, Wenqing Chu, Ye Wu, Weihang Yuan, Fanglong Liu, Qi Zhang, Fu Li, Haocheng Feng, Errui Ding, and Jingdong Wang.
\newblock Videogen: A reference-guided latent diffusion approach for high definition text-to-video generation.
\newblock \emph{arXiv preprint arXiv:2309.00398}, 2023{\natexlab{c}}.

\bibitem[Lin et~al.(2014)Lin, Maire, Belongie, Hays, Perona, Ramanan, Doll{\'{a}}r, and Zitnick]{LinMBHPRDZ14}
Tsung{-}Yi Lin, Michael Maire, Serge~J. Belongie, James Hays, Pietro Perona, Deva Ramanan, Piotr Doll{\'{a}}r, and C.~Lawrence Zitnick.
\newblock Microsoft {COCO:} common objects in context.
\newblock In \emph{Proceedings of the ECCV}, pages 740--755, 2014.

\bibitem[Liu et~al.(2023{\natexlab{a}})Liu, Chen, Yuan, Mei, Liu, Mandic, Wang, and Plumbley]{LiuCYMLM0P23}
Haohe Liu, Zehua Chen, Yi Yuan, Xinhao Mei, Xubo Liu, Danilo~P. Mandic, Wenwu Wang, and Mark~D. Plumbley.
\newblock Audioldm: Text-to-audio generation with latent diffusion models.
\newblock In \emph{ICML}, 2023{\natexlab{a}}.

\bibitem[Liu et~al.(2023{\natexlab{b}})Liu, Li, Li, and Lee]{liu2023improved}
Haotian Liu, Chunyuan Li, Yuheng Li, and Yong~Jae Lee.
\newblock Improved baselines with visual instruction tuning.
\newblock In \emph{arXiv}, 2023{\natexlab{b}}.

\bibitem[Liu et~al.(2023{\natexlab{c}})Liu, Li, Wu, and Lee]{liu2023visual}
Haotian Liu, Chunyuan Li, Qingyang Wu, and Yong~Jae Lee.
\newblock Visual instruction tuning.
\newblock In \emph{arXiv}, 2023{\natexlab{c}}.

\bibitem[Liu et~al.(2023{\natexlab{d}})Liu, Zeng, Ren, Li, Zhang, Yang, Jiang, Li, Yang, Su, et~al.]{liu2023grounding}
Shilong Liu, Zhaoyang Zeng, Tianhe Ren, Feng Li, Hao Zhang, Jie Yang, Qing Jiang, Chunyuan Li, Jianwei Yang, Hang Su, et~al.
\newblock Grounding dino: Marrying dino with grounded pre-training for open-set object detection.
\newblock In \emph{arXiv}, 2023{\natexlab{d}}.

\bibitem[Meng et~al.(2022)Meng, He, Song, Song, Wu, Zhu, and Ermon]{MengHSSWZE22}
Chenlin Meng, Yutong He, Yang Song, Jiaming Song, Jiajun Wu, Jun{-}Yan Zhu, and Stefano Ermon.
\newblock Sdedit: Guided image synthesis and editing with stochastic differential equations.
\newblock In \emph{ICLR}, 2022.

\bibitem[Michaels et~al.(2024)Michaels, Li, Yao, Yu, Wood-Doughty, and Metze]{michaels2024audio}
Jackson Michaels, Juncheng~B Li, Laura Yao, Lijun Yu, Zach Wood-Doughty, and Florian Metze.
\newblock Audio-journey: Open domain latent diffusion based text-to-audio generation.
\newblock In \emph{ICASSP 2024-2024 IEEE International Conference on Acoustics, Speech and Signal Processing (ICASSP)}, pages 6960--6964. IEEE, 2024.

\bibitem[Nichol et~al.(2022)Nichol, Dhariwal, Ramesh, Shyam, Mishkin, McGrew, Sutskever, and Chen]{NicholDRSMMSC22}
Alexander~Quinn Nichol, Prafulla Dhariwal, Aditya Ramesh, Pranav Shyam, Pamela Mishkin, Bob McGrew, Ilya Sutskever, and Mark Chen.
\newblock {GLIDE:} towards photorealistic image generation and editing with text-guided diffusion models.
\newblock In \emph{Proceedings of the ICML}, pages 16784--16804, 2022.

\bibitem[OpenAI(2022)]{chatgpt}
OpenAI.
\newblock Introducing chatgpt.
\newblock 2022.

\bibitem[Perazzi et~al.(2016)Perazzi, Pont{-}Tuset, McWilliams, Gool, Gross, and Sorkine{-}Hornung]{PerazziPMGGS16}
Federico Perazzi, Jordi Pont{-}Tuset, Brian McWilliams, Luc~Van Gool, Markus~H. Gross, and Alexander Sorkine{-}Hornung.
\newblock A benchmark dataset and evaluation methodology for video object segmentation.
\newblock In \emph{Proceedings of the CVPR}, pages 724--732, 2016.

\bibitem[Qu et~al.(2023)Qu, Wu, Fei, Nie, and Chua]{qu2023layoutllm}
Leigang Qu, Shengqiong Wu, Hao Fei, Liqiang Nie, and Tat-Seng Chua.
\newblock Layoutllm-t2i: Eliciting layout guidance from llm for text-to-image generation.
\newblock In \emph{Proceedings of the 31st ACM International Conference on Multimedia}, pages 643--654, 2023.

\bibitem[Qu et~al.(2024)Qu, Wang, Li, Zhang, Nie, and Chua]{qu2024discriminative}
Leigang Qu, Wenjie Wang, Yongqi Li, Hanwang Zhang, Liqiang Nie, and Tat-Seng Chua.
\newblock Discriminative probing and tuning for text-to-image generation.
\newblock In \emph{Proceedings of the IEEE/CVF Conference on Computer Vision and Pattern Recognition}, pages 7434--7444, 2024.

\bibitem[Radford et~al.(2019)Radford, Wu, Child, Luan, Amodei, Sutskever, et~al.]{radford2019language}
Alec Radford, Jeffrey Wu, Rewon Child, David Luan, Dario Amodei, Ilya Sutskever, et~al.
\newblock Language models are unsupervised multitask learners.
\newblock In \emph{OpenAI blog}, 2019.

\bibitem[Rombach et~al.(2022)Rombach, Blattmann, Lorenz, Esser, and Ommer]{robin2022high}
Robin Rombach, Andreas Blattmann, Dominik Lorenz, Patrick Esser, and Bj{\"{o}}rn Ommer.
\newblock High-resolution image synthesis with latent diffusion models.
\newblock In \emph{CVPR}, 2022.

\bibitem[Shanahan(2022)]{Shanahan2022talking}
Murray Shanahan.
\newblock Talking about large language models.
\newblock In \emph{arXiv}, 2022.

\bibitem[Sharma et~al.(2018)Sharma, Ding, Goodman, and Soricut]{SoricutDSG18}
Piyush Sharma, Nan Ding, Sebastian Goodman, and Radu Soricut.
\newblock Conceptual captions: {A} cleaned, hypernymed, image alt-text dataset for automatic image captioning.
\newblock In \emph{Proceedings of the ACL}, pages 2556--2565, 2018.

\bibitem[Shen et~al.(2023)Shen, Song, Tan, Li, Lu, and Zhuang]{shen2023hugginggpt}
Yongliang Shen, Kaitao Song, Xu Tan, Dongsheng Li, Weiming Lu, and Yueting Zhuang.
\newblock Hugginggpt: Solving ai tasks with chatgpt and its friends in huggingface.
\newblock In \emph{arXiv}, 2023.

\bibitem[Singer et~al.(2022)Singer, Polyak, Hayes, Yin, An, Zhang, Hu, Yang, Ashual, Gafni, Parikh, Gupta, and Taigman]{abs-2209-14792}
Uriel Singer, Adam Polyak, Thomas Hayes, Xi Yin, Jie An, Songyang Zhang, Qiyuan Hu, Harry Yang, Oron Ashual, Oran Gafni, Devi Parikh, Sonal Gupta, and Yaniv Taigman.
\newblock Make-a-video: Text-to-video generation without text-video data.
\newblock In \emph{arXiv}, 2022.

\bibitem[Soomro et~al.(2012)Soomro, Zamir, and Shah]{soomro2012ucf101}
Khurram Soomro, Amir~Roshan Zamir, and Mubarak Shah.
\newblock Ucf101: A dataset of 101 human actions classes from videos in the wild.
\newblock \emph{arXiv preprint arXiv:1212.0402}, 2012.

\bibitem[Su et~al.(2022)Su, Lan, Liu, Liu, Yogatama, Wang, Kong, and Collier]{su2022language}
Yixuan Su, Tian Lan, Yahui Liu, Fangyu Liu, Dani Yogatama, Yan Wang, Lingpeng Kong, and Nigel Collier.
\newblock Language models can see: Plugging visual controls in text generation.
\newblock In \emph{arXiv}, 2022.

\bibitem[Su et~al.(2023)Su, Lan, Li, Xu, Wang, and Cai]{su2023pandagpt}
Yixuan Su, Tian Lan, Huayang Li, Jialu Xu, Yan Wang, and Deng Cai.
\newblock Pandagpt: One model to instruction-follow them all.
\newblock In \emph{arXiv}, 2023.

\bibitem[Tang et~al.(2024)Tang, Yang, Zhu, Zeng, and Bansal]{abs-2305-11846}
Zineng Tang, Ziyi Yang, Chenguang Zhu, Michael Zeng, and Mohit Bansal.
\newblock Any-to-any generation via composable diffusion.
\newblock In \emph{NeurIPS}, 2024.

\bibitem[Taylor et~al.(2022)Taylor, Kardas, Cucurull, Scialom, Hartshorn, Saravia, Poulton, Kerkez, and Stojnic]{taylor2022galactica}
Ross Taylor, Marcin Kardas, Guillem Cucurull, Thomas Scialom, Anthony Hartshorn, Elvis Saravia, Andrew Poulton, Viktor Kerkez, and Robert Stojnic.
\newblock Galactica: A large language model for science.
\newblock In \emph{arXiv}, 2022.

\bibitem[Team et~al.(2023)Team, Anil, Borgeaud, Wu, Alayrac, Yu, Soricut, Schalkwyk, Dai, Hauth, et~al.]{team2023gemini}
Gemini Team, Rohan Anil, Sebastian Borgeaud, Yonghui Wu, Jean-Baptiste Alayrac, Jiahui Yu, Radu Soricut, Johan Schalkwyk, Andrew~M Dai, Anja Hauth, et~al.
\newblock Gemini: a family of highly capable multimodal models.
\newblock In \emph{arXiv}, 2023.

\bibitem[Touvron et~al.(2023{\natexlab{a}})Touvron, Lavril, Izacard, Martinet, Lachaux, Lacroix, Rozi{\`e}re, Goyal, Hambro, Azhar, et~al.]{touvron2023llama}
Hugo Touvron, Thibaut Lavril, Gautier Izacard, Xavier Martinet, Marie-Anne Lachaux, Timoth{\'e}e Lacroix, Baptiste Rozi{\`e}re, Naman Goyal, Eric Hambro, Faisal Azhar, et~al.
\newblock Llama: Open and efficient foundation language models.
\newblock In \emph{arXiv}, 2023{\natexlab{a}}.

\bibitem[Touvron et~al.(2023{\natexlab{b}})Touvron, Martin, Stone, Albert, Almahairi, Babaei, Bashlykov, Batra, Bhargava, Bhosale, et~al.]{touvron2023llama2}
Hugo Touvron, Louis Martin, Kevin Stone, Peter Albert, Amjad Almahairi, Yasmine Babaei, Nikolay Bashlykov, Soumya Batra, Prajjwal Bhargava, Shruti Bhosale, et~al.
\newblock Llama 2: Open foundation and fine-tuned chat models.
\newblock In \emph{arXiv}, 2023{\natexlab{b}}.

\bibitem[Veaux et~al.(2017)Veaux, Yamagishi, MacDonald, et~al.]{veaux2017cstr}
Christophe Veaux, Junichi Yamagishi, Kirsten MacDonald, et~al.
\newblock Cstr vctk corpus: English multi-speaker corpus for cstr voice cloning toolkit.
\newblock \emph{CSTR}, 6:\penalty0 15, 2017.

\bibitem[Wang et~al.(2022{\natexlab{a}})Wang, Yang, Hu, Li, Lin, Gan, Liu, Liu, and Wang]{WangYHLLGLLW22}
Jianfeng Wang, Zhengyuan Yang, Xiaowei Hu, Linjie Li, Kevin Lin, Zhe Gan, Zicheng Liu, Ce Liu, and Lijuan Wang.
\newblock {GIT:} {A} generative image-to-text transformer for vision and language.
\newblock \emph{Trans. Mach. Learn. Res.}, 2022, 2022{\natexlab{a}}.

\bibitem[Wang et~al.(2022{\natexlab{b}})Wang, Yang, Men, Lin, Bai, Li, Ma, Zhou, Zhou, and Yang]{wang2022ofa}
Peng Wang, An Yang, Rui Men, Junyang Lin, Shuai Bai, Zhikang Li, Jianxin Ma, Chang Zhou, Jingren Zhou, and Hongxia Yang.
\newblock Ofa: Unifying architectures, tasks, and modalities through a simple sequence-to-sequence learning framework.
\newblock In \emph{ICML}, 2022{\natexlab{b}}.

\bibitem[Wang et~al.(2022{\natexlab{c}})Wang, Yi, Fu, Tao, and Wen]{WangYFTW22}
Tao Wang, Jiangyan Yi, Ruibo Fu, Jianhua Tao, and Zhengqi Wen.
\newblock Campnet: Context-aware mask prediction for end-to-end text-based speech editing.
\newblock \emph{{IEEE} {ACM} Trans. Audio Speech Lang. Process.}, 30:\penalty0 2241--2254, 2022{\natexlab{c}}.

\bibitem[Wu et~al.(2023{\natexlab{a}})Wu, Yin, Qi, Wang, Tang, and Duan]{wu2023visual}
Chenfei Wu, Shengming Yin, Weizhen Qi, Xiaodong Wang, Zecheng Tang, and Nan Duan.
\newblock Visual chatgpt: Talking, drawing and editing with visual foundation models.
\newblock In \emph{arXiv}, 2023{\natexlab{a}}.

\bibitem[Wu et~al.(2023{\natexlab{b}})Wu, Ge, Wang, Lei, Gu, Hsu, Shan, Qie, and Shou]{abs-2212-11565}
Jay~Zhangjie Wu, Yixiao Ge, Xintao Wang, Weixian Lei, Yuchao Gu, Wynne Hsu, Ying Shan, Xiaohu Qie, and Mike~Zheng Shou.
\newblock Tune-a-video: One-shot tuning of image diffusion models for text-to-video generation.
\newblock In \emph{ICCV}, 2023{\natexlab{b}}.

\bibitem[Wu et~al.(2024)Wu, Fei, Qu, Ji, and Chua]{wu2023next}
Shengqiong Wu, Hao Fei, Leigang Qu, Wei Ji, and Tat-Seng Chua.
\newblock Next-gpt: Any-to-any multimodal llm.
\newblock In \emph{ICML}, 2024.

\bibitem[Xu et~al.(2023)Xu, Ye, Yan, Shi, Ye, Xu, Li, Bi, Qian, Wang, Xu, Zhang, Huang, Huang, and Zhou]{XuYYSYXLBQWXZH023}
Haiyang Xu, Qinghao Ye, Ming Yan, Yaya Shi, Jiabo Ye, Yuanhong Xu, Chenliang Li, Bin Bi, Qi Qian, Wei Wang, Guohai Xu, Ji Zhang, Songfang Huang, Fei Huang, and Jingren Zhou.
\newblock mplug-2: {A} modularized multi-modal foundation model across text, image and video.
\newblock In \emph{Proceedings of the ICML}, pages 38728--38748, 2023.

\bibitem[Xu et~al.(2016)Xu, Mei, Yao, and Rui]{XuMYR16}
Jun Xu, Tao Mei, Ting Yao, and Yong Rui.
\newblock {MSR-VTT:} {A} large video description dataset for bridging video and language.
\newblock In \emph{Proceedings of the CVPR}, pages 5288--5296, 2016.

\bibitem[Yang et~al.(2023)Yang, Yu, Wang, Wang, Weng, Zou, and Yu]{YangYWWWZY23}
Dongchao Yang, Jianwei Yu, Helin Wang, Wen Wang, Chao Weng, Yuexian Zou, and Dong Yu.
\newblock Diffsound: Discrete diffusion model for text-to-sound generation.
\newblock \emph{{IEEE} {ACM} Trans. Audio Speech Lang. Process.}, 31:\penalty0 1720--1733, 2023.

\bibitem[Zhang et~al.(2023{\natexlab{a}})Zhang, Li, Zhang, Zhan, Wang, Zhou, and Qiu]{zhang2023speechgpt}
Dong Zhang, Shimin Li, Xin Zhang, Jun Zhan, Pengyu Wang, Yaqian Zhou, and Xipeng Qiu.
\newblock Speechgpt: Empowering large language models with intrinsic cross-modal conversational abilities.
\newblock In \emph{arXiv}, 2023{\natexlab{a}}.

\bibitem[Zhang et~al.(2023{\natexlab{b}})Zhang, Li, and Bing]{zhang2023video}
Hang Zhang, Xin Li, and Lidong Bing.
\newblock Video-llama: An instruction-tuned audio-visual language model for video understanding.
\newblock In \emph{arXiv}, 2023{\natexlab{b}}.

\bibitem[Zhang et~al.(2020)Zhang, Shi, Yuan, Li, Wang, Hu, and Zha]{ZhangSY0WHZ20}
Ziqi Zhang, Yaya Shi, Chunfeng Yuan, Bing Li, Peijin Wang, Weiming Hu, and Zheng{-}Jun Zha.
\newblock Object relational graph with teacher-recommended learning for video captioning.
\newblock In \emph{Proceedings of the CVPR}, pages 13275--13285, 2020.

\bibitem[Zhu et~al.(2023)Zhu, Chen, Shen, Li, and Elhoseiny]{zhu2023minigpt}
Deyao Zhu, Jun Chen, Xiaoqian Shen, Xiang Li, and Mohamed Elhoseiny.
\newblock Minigpt-4: Enhancing vision-language understanding with advanced large language models.
\newblock In \emph{arXiv}, 2023.

\end{thebibliography}
}
\clearpage
\setcounter{page}{1}
\maketitlesupplementary
\renewcommand{\thesection}{\Alph{section}}
\setcounter{section}{0}

\section{Definition of AMMG Task}
We give a formal problem definition of Any-to-Many Modalities Generation (AMMG).
Let $\mathcal{X} = \{\text{Text}, X_1, X_2, \dots, X_n\}$ represent the set of available modalities, where each $X_i$ is a unique modality (e.g., image, audio, video, box, mask).
For simplicity, we may alternatively use $X$ to represent any modality $X_i$.
\textbf{Input Query}:  The input query $Q$ is defined as either a single text input or a combination of text with one additional modality: $Q = \{\text{Text}\} d \text{or} d Q = \{\text{Text}, X_i\}$.
\textbf{Output Generation}: The output $Y$ is defined as a sequence containing any number of modalities from $\mathcal{X}$, structured as: $Y = (y_1, y_2, \dots, y_k), \, \text{where each } y_i \in \mathcal{X} d \text{and} d k \geq 1$. This allows $Y$ to be any arbitrary combination of modalities in response to $Q$.
\textbf{Objective}: Given an input query $Q$, the objective is to generate a multimodal output $Y$ that accurately satisfies the instructional requirements of $Q$, integrating all requested modalities in a single response, avoiding multiple rounds of interaction to satisfy multimodal requirements.

\section{Motivation of AMMG Task}
\label{sec:motivation_ammg}
\begin{figure}[!ht]
\vspace{-1mm}
\centering
\includegraphics[width=0.95\columnwidth]{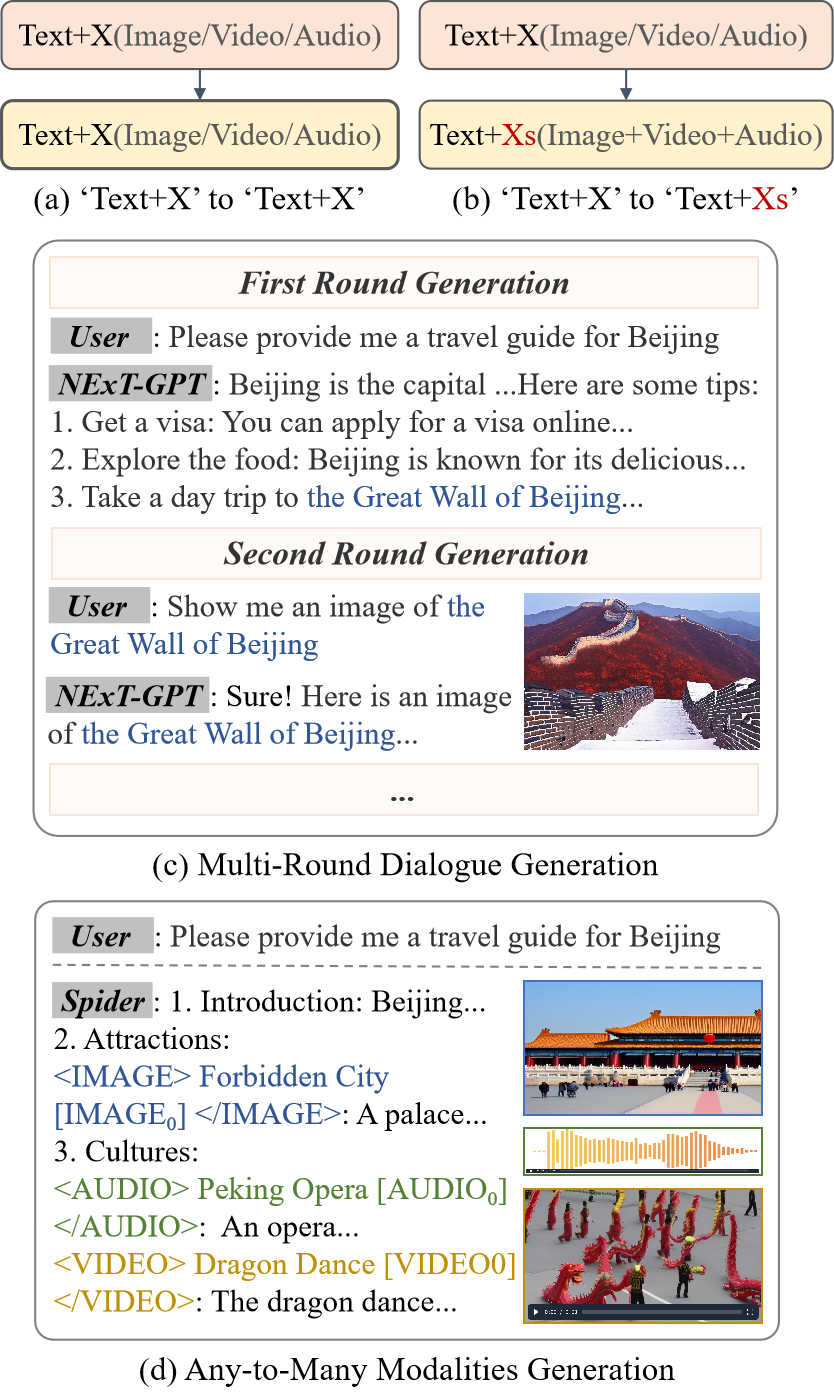}
\caption{Comparison between (a) the \textbf{X-to-X (Any-to-Any) MLLMs} support the input and output of pairwise modalities '\textbf{Text + X}', and (b) our \textbf{X-to-Xs (Any-to-Many) Spider model} produces many modalities '\textbf{Text + Xs}'. X denotes any-one-modality such as one of image or video or audio, and Xs means arbitrary-combination-modalities such as the combination of image and video and audio. 
(c) Multi-Round Dialogue Generation. (d) Any-to-Many Modalities Generation.}
\vspace{-2mm}
\label{fig:motivation_xs}
\end{figure}

As shown in Fig.~\ref{fig:motivation_xs}(a), the \textbf{X-to-X (Any-to-Any) MLLMs} are restricted to generate pairwise modalities '\textbf{Text + X}' within a single interaction, such as 'Text + Image' or 'Text + Audio'. 
For example, in Fig.~\ref{fig:motivation_xs}(c), when a user asks "Please provide me a travel guide for Beijing", the model first responds with the text output, i.e., 'Text' to 'Text'. In subsequent interaction, the user needs to further request "Show me an image of the Great Wall of Beijing", to get the required image, i.e., 'Text' to 'Text + Image'.
These MLLMs based on \textit{Multi-Round Dialogue Generation} paradigm, require several rounds of user questions and do not allow for a seamless integration of multiple modalities within a single interaction. Each pair of modalities is handled independently, resulting in a fragmented user experience where the responses feel disjointed rather than cohesive.

In contrast, as illustrated in Fig.~\ref{fig:motivation_xs}(b), our proposed \textbf{X-to-Xs (Any-to-Many) Spider model}, achieves \textbf{Any-to-Many Modalities Generation (AMMG)} in a single response. For the user question "Please provide me a travel guide for Beijing", Spider generates a cohesive output that combines text, image, audio, and video in a single response, greatly enhancing the user experience by providing comprehensive many-modal content all at once.

\section{Related Work}
\subsection{Large Language Models}
Large Language Models (LLMs), such as BERT \cite{Devlin2019BERT}, GPT-2 \cite{radford2019language}, GPT-3 \cite{brown2020language}, PaLM \cite{chowdhery2022palm}, Galactica \cite{taylor2022galactica}, and LLaMA \cite{touvron2023llama}, are Transformer-based models with hundreds of billions of parameters, trained on vast text datasets \cite{Shanahan2022talking}. These models excel in understanding natural language and performing complex tasks, especially text generation. ChatGPT \cite{chatgpt}, powered by GPT models, demonstrates the conversational capabilities of LLMs, contributing to the growing interest in artificial general intelligence (AGI). The rapid development of LLMs is reshaping AI research, with LLMs now seen as a versatile tool for a wide range of language-related tasks.

\subsection{Large Multimodal Models}
To build foundational Multimodal LLMs (MLLMs), researchers align pre-trained encoders from various modalities with the textual space of LLMs, enabling them to process multimodal inputs \cite{huang2023language, zhu2023minigpt, su2022language, koh2023generating, alayrac2022flamingo, li2023blip, liu2023visual}. For example, Flamingo \cite{alayrac2022flamingo} connects a fixed image encoder to LLMs using cross-attention, while LLaVA \cite{liu2023visual} links image and word spaces via projection. BLIP-2 \cite{li2023blip} uses a Q-Former to translate image queries into LLMs. Similar approaches are applied to videos (e.g., Video-Chat \cite{li2023videochat}, Video-LLaMA \cite{zhang2023video}) and audios (e.g., SpeechGPT \cite{zhang2023speechgpt}). PandaGPT \cite{su2023pandagpt} extends this to six modalities using ImageBind \cite{girdhar2023imagebind}.

However, existing MLLMs only perceive multimodal data and cannot generate content in arbitrary modalities. To address this, approaches like Visual-ChatGPT \cite{wu2023visual}, HuggingGPT \cite{shen2023hugginggpt}, and AudioGPT \cite{huang2023audiogpt} use LLMs as decision-makers, incorporating external multimodal encoders and decoders for multimodal input-output. Despite this, discrete text-message-based pipelines can introduce noise and hinder semantic understanding. NExT-GPT \cite{wu2023next} overcomes this by learning an end-to-end multimodal input-output LLM, capable of handling any combination of text, image, video, and audio.

\begin{figure}[t]
%\vspace{-2mm}
\centering
\includegraphics[width=1.0\columnwidth]{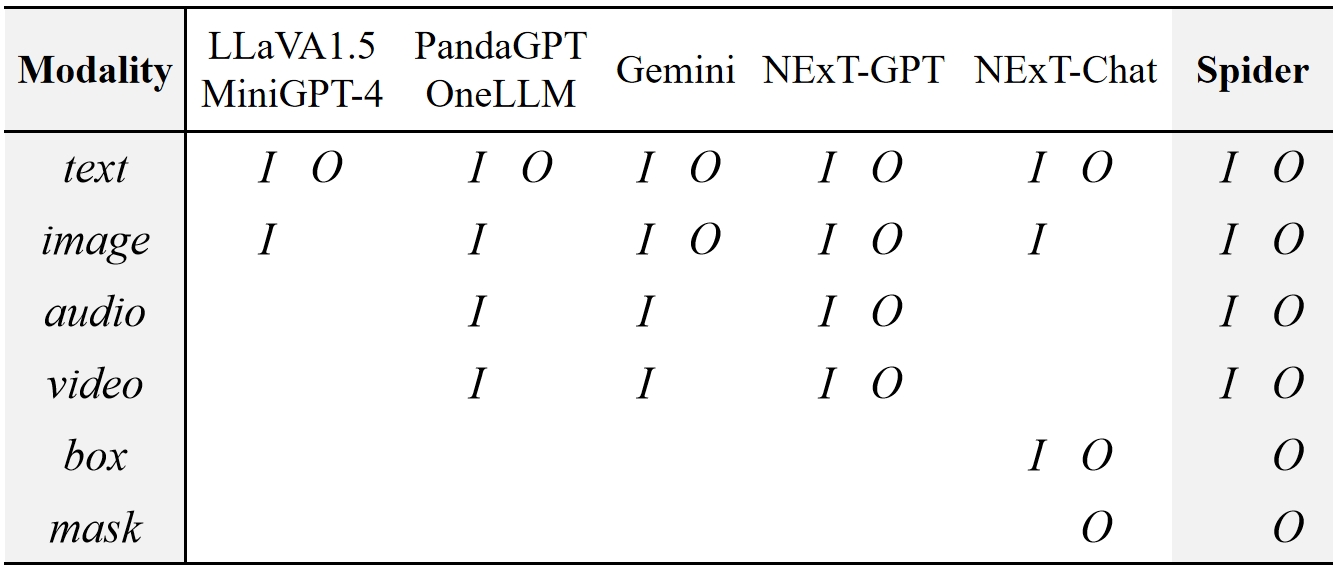}
\caption{Our \textbf{Spider} supports many modalities for \textit{I} and \textit{O} (i.e., input and output) including text, image, audio, video, box, mask.}
\vspace{-2mm}
\label{fig:supported_x}
\end{figure}

However, these X-to-X MLLMs are limited to generating pairwise modalities 'Text + X' within a single interaction.
In contrast, our proposed X-to-Xs Spider model aims for Any-to-Many Modalities Generation in a single response, supporting arbitrary combinations of a wider range of modalities as shown in Fig.~\ref{fig:supported_x}, including text, image, audio, video, box, and mask.

%\section{Efficiency of Unified Decoder Projector}
%Comparison between current MLLMs with Multiple Projectors for LLM-Decoders alignment, and our Unified Decoder Projector, where \{M-Query\} denotes the list of learnable Modality Query (M-Query) for the corresponding output modalities. The parameter of \{M-Query\} is far less than Projector. As the number of modalities increases, the number of Projectors (i.e., the parameters) in Multiple Projectors grows linearly. However, our Unified Decoder Projector adds the modality query in \{M-Query\}, with merely parameter growth.

\begin{table*}[t]
\begin{minipage}{\textwidth}
\centering
\begin{minipage}[b]{0.3\textwidth}
% \centering
\fontsize{8}{11}\selectfont
\setlength{\tabcolsep}{1.2mm}
\centering
\begin{tabular}{lccc}
\hline
\bf Method & \bf B@4 & \bf METEOR & \bf CIDEr \\
\hline
Oscar \cite{Li0LZHZWH0WCG20}  &  36.58  & 30.4  & 124.12 \\
BLIP-2 \cite{li2023blip} &  43.7 &  --- &  145.8 \\
OFA \cite{wang2022ofa} &  44.9 &  32.5 &  154.9 \\
CoDi \cite{abs-2305-11846} &   40.2 &  31.0 &  149.9 \\
\cdashline{1-4}
NExT-GPT \cite{wu2023next} & 45.1   & 34.1  & 158.3  \\
\bf Our Spider & \bf 45.9   & \bf 34.3  & \bf 158.8  \\
\hline
\end{tabular}
%\vspace{1mm}
\captionof{table}{
\label{tab:I2T-res}
Image-to-Text generation on COCO-caption \cite{LinMBHPRDZ14}.
} 
\end{minipage}
\hfill
\begin{minipage}[b]{0.3\textwidth}
\centering
\fontsize{8}{11}\selectfont
\setlength{\tabcolsep}{1.2mm}
\begin{tabular}{lcc}
\hline
% \cmidrule(r){2-3}\cmidrule(r){4-5}
\bf Method & \bf SPIDEr ($\uparrow$) & \bf CIDEr ($\uparrow$) \\
\hline
AudioCaps \cite{KimKLK19}&	 0.369& 0.593 \\   
BART \cite{GontierSC21} &  0.465& 0.753 \\   
AL-MixGen \cite{abs-2210-17143} &  0.466 & 0.755 \\   
CoDi \cite{abs-2305-11846} & 0.480 & 0.789 \\
\cdashline{1-3}
NExT-GPT \cite{wu2023next} &   0.534 &   0.807 \\    
\bf Our Spider &  \bf  0.537 & \bf  0.819 \\ 
\hline
\end{tabular}
%\vspace{1mm}
\captionof{table}{
\label{tab:A2T-res}
Audio-to-Text generation on AudioCaps \cite{KimKLK19}.
}
\end{minipage}
\hfill
\begin{minipage}[b]{0.30\textwidth}
\centering
\fontsize{8}{11}\selectfont
\setlength{\tabcolsep}{1.1mm}
\begin{tabular}{lcc}
\hline
% \cmidrule(r){2-3}\cmidrule(r){4-5}
\bf Method & \bf B@4 ($\uparrow$) & \bf METEOR ($\uparrow$)  \\
\hline
ORG-TRL \cite{ZhangSY0WHZ20}&	43.6 &  28.8 \\   
GIT \cite{WangYHLLGLLW22} & 	 54.8 &  33.1 \\   
mPLUG-2 \cite{XuYYSYXLBQWXZH023} & 57.8 &  34.9 \\  
CoDi \cite{abs-2305-11846} & 52.1 &  32.5 \\
\cdashline{1-3}
NExT-GPT \cite{wu2023next} &  58.8 &  39.6 \\    
\bf Our Spider & \bf 59.8 &  \bf 40.2 \\   
\hline
\end{tabular}
%\vspace{1mm}
\captionof{table}{
\label{tab:V2T-res}
Video-to-Text generation on MSR-VTT \cite{XuMYR16}.
}
\end{minipage}
\vspace{-2mm}
\end{minipage}
% \vspace{-4mm}
\end{table*}

\begin{table*}[t]
\begin{minipage}{\textwidth}
\centering
\begin{minipage}[b]{0.3\textwidth}
% \centering
\fontsize{8}{11}\selectfont
\setlength{\tabcolsep}{2.5mm}
\centering
\begin{tabular}{lc}
\hline
\bf Method & \bf FID ($\downarrow$) \\
\hline
CogVideo \cite{DingYHZZYLZSYT21}& 	27.10 \\  
GLIDE \cite{NicholDRSMMSC22} & 	12.24 \\
CoDi \cite{abs-2305-11846} &  11.26 \\
SD \cite{robin2022high} & 11.21 \\
\cdashline{1-2}
NExT-GPT \cite{wu2023next} & 11.18\\
\bf Our Spider & \bf 11.13\\
\hline
\end{tabular}
%\vspace{1mm}
\captionof{table}{
\label{tab:T2I-res}
Text-to-Image generation on COCO-caption \cite{LinMBHPRDZ14}.
}
\end{minipage}
\hfill
\begin{minipage}[b]{0.3\textwidth}
\centering
\fontsize{8}{11}\selectfont
\setlength{\tabcolsep}{1.5mm}
\begin{tabular}{lcc}
\hline
% \cmidrule(r){2-3}\cmidrule(r){4-5}
\bf Method & \bf FD ($\downarrow$) & \bf IS ($\uparrow$) \\
\hline
DiffSound \cite{YangYWWWZY23}&	47.68 & 4.01 \\   
AudioLDM-S \cite{LiuCYMLM0P23} & 29.48 & 6.90 \\   
AudioLDM-L \cite{LiuCYMLM0P23} &  23.31  &  8.13 \\  
CoDi \cite{abs-2305-11846} & \bf 22.90 & 8.77 \\
\cdashline{1-3}
NExT-GPT \cite{wu2023next} & 23.25 & 8.67 \\    
\bf Our Spider & 23.02 & \bf 8.84 \\   
\hline
\end{tabular}
%\vspace{1mm}
\captionof{table}{
\label{tab:T2A-res}
Text-to-Audio generation on AudioCaps \cite{KimKLK19}.
}
\end{minipage}
\hfill
\begin{minipage}[b]{0.3\textwidth}
\centering
\fontsize{8}{11}\selectfont
\setlength{\tabcolsep}{1.1mm}
\begin{tabular}{lcc}
\hline
% \cmidrule(r){2-3}\cmidrule(r){4-5}
\bf Method & \bf FID ($\downarrow$) & \bf CLIPSIM ($\uparrow$) \\
\hline
CogVideo \cite{abs-2205-15868}&	23.59 & 	0.2631 \\   
MakeVideo \cite{abs-2209-14792} & 	13.17 & 	0.3049 \\   
Latent-VDM \cite{robin2022high} &  14.25 & 	0.2756 \\   
Latent-Shift \cite{Jie2023Latent} & 15.23 & 	0.2773 \\
%CoDi \cite{abs-2305-11846} & --- &  0.2890 \\
\cdashline{1-3}
NExT-GPT \cite{wu2023next} &  12.69 &  0.3197 \\    
\bf Our Spider & \bf 12.62 &  \bf 0.3258 \\ 
\hline
\end{tabular}
%\vspace{1mm}
\captionof{table}{
\label{tab:T2V-res}
Text-to-Video generation on MSR-VTT \cite{XuMYR16}.
}
\end{minipage}
% \vspace{-2mm}
\end{minipage}
\vspace{-2mm}
\end{table*}

\begin{table*}[!t]
\begin{minipage}{\textwidth}
\centering
\begin{minipage}[b]{0.3\textwidth}
% \centering
\fontsize{8}{11}\selectfont
\setlength{\tabcolsep}{1.7mm}
\centering
\begin{tabular}{lcc}
\hline
\bf Method & \bf CLIP ($\uparrow$) & \bf FID ($\downarrow$) \\
\hline
BLDM \cite{AvrahamiFL23} &  29.95 &  6.14 \\
DiffEdit \cite{CouaironVSC23} &   29.30 &\bf 3.78 \\
PFB-Diff \cite{abs-2306-16894} &\bf  30.81 &  5.93 \\
\cdashline{1-3}
NExT-GPT \cite{wu2023next} & 29.32   & 6.62  \\
\bf Our Spider & 30.52   & 5.33  \\
\hline
\end{tabular}
%\vspace{1mm}
\captionof{table}{
\label{tab:TI2I-res}
Image-to-Image generation (text-conditioned image editing for object) on COCO \cite{LinMBHPRDZ14}.
} 
\end{minipage}
\hfill
\begin{minipage}[b]{0.3\textwidth}
\centering
\fontsize{8}{11}\selectfont
\setlength{\tabcolsep}{1.7mm}
\begin{tabular}{lc}
\hline
\bf Method & \bf MCD ($\downarrow$) \\
\hline
CampNet \cite{WangYFTW22}& 0.380 \\ 
MakeAudio \cite{HuangHY0LLYLYZ23}  & 0.375 \\   
AudioLDM-L \cite{LiuCYMLM0P23}  & 0.349 \\
\cdashline{1-2}
NExT-GPT \cite{wu2023next} &  0.300 \\    
\bf Our Spider &  \bf 0.279 \\    
\hline
\end{tabular}
%\vspace{1mm}
\captionof{table}{
\label{tab:TA2A-res}
Audio-to-Audio generation (text-conditioned speech editing) on VCTK \cite{veaux2017cstr}.
}
\end{minipage}
\hfill
\begin{minipage}[b]{0.3\textwidth}
\centering
\fontsize{8}{11}\selectfont
\setlength{\tabcolsep}{0.8mm}
\begin{tabular}{lcc}
\hline
\bf Method & \bf CLIP-T ($\uparrow$) & \bf CLIP-I ($\uparrow$)  \\
\hline  
TuneVideo \cite{abs-2212-11565} & 	  0.2758 & 0.9240 \\   
SDEdit \cite{MengHSSWZE22} &  0.2775 & 0.8731 \\   
Pix2Video \cite{abs-2303-12688} & \bf 0.2891  & \bf 0.9767 \\
\cdashline{1-3}
NExT-GPT \cite{wu2023next} & 0.2684 &  0.9647 \\    
\bf Our Spider & 0.2782 &  0.9715 \\ 
\hline
\end{tabular}
%\vspace{1mm}
\captionof{table}{
\label{tab:TV2V-res}
Video-to-Video generation (text-conditioned video editing) on DAVIS \cite{PerazziPMGGS16}.
}
\end{minipage}
% \vspace{-2mm}
\end{minipage}
\vspace{-2mm}
\end{table*}

\begin{table*}[ht!]
\centering
\setlength{\abovecaptionskip}{0.15cm}
\resizebox{0.95\textwidth}{!}
{
\setlength\tabcolsep{7pt}
\renewcommand\arraystretch{1.1}
\begin{tabular}{p{3cm}<{\raggedleft}|ccc|ccc|ccc}
\hline
\multicolumn{1}{c|}{\multirow{2}{*}{\bf Method}} & \multicolumn{3}{c|}{\bf COCO-NSS1K (in-distribution)}  & \multicolumn{3}{c|}{\bf CC-500 (out-of-distribution) } & \multicolumn{3}{c}{\bf ABC-6K (mixed-distribution)} \\ 
\cline{2-10}
 & CLIP & BLIP-M  & BLIP-C& CLIP & BLIP-M & BLIP-C  & CLIP & BLIP-M & BLIP-C \\
\hline
\multicolumn{1}{l|}{SD \cite{robin2022high}}    & 33.27  & 67.96  & 39.48  & 34.82  & 70.95  & 40.36   & 35.33  & 72.03  & 40.82  \\
\multicolumn{1}{l|}{{StructureDiffusion}~\cite{feng2022training}} & -     & -     & -     & 33.71  & 66.71  & 39.54  &   34.95 & 69.55 & 40.69 \\
\multicolumn{1}{l|}{{HN-DiffusionITM}~\cite{krojer2023diffusion}}    & 33.26  & 70.06  & \textbf{40.14}  & 34.15  & 68.77  & 40.30  & 35.02  & 72.28  & 41.12   \\
\multicolumn{1}{l|} {DPT~\cite{qu2024discriminative}} & \textbf{33.85}  & \textbf{71.84}  & 40.11  & \textbf{35.97}  & \textbf{76.74}  & \textbf{41.15}    & \textbf{35.88}  & \textbf{75.88}  & \textbf{41.26} \\
\hdashline
\multicolumn{1}{l|}{\bf Our Spider} & 33.40  & 68.23  & 39.55  & 34.98 & 71.12  & 40.48  & 35.37 &  72.30 & 40.91\\
\hline
\end{tabular}
}
\caption{Text-to-Image generation on COCO-NSS1K~\cite{qu2023layoutllm}, CC-500~\cite{feng2022training}, and ABC-6K~\cite{feng2022training}.}
\label{tab:perf_comp_gen}
\end{table*}

\begin{table}[htbp]
\small
\centering
\begin{tabular}{l|cccc}
\hline
\bf Method  & \bf FD ($\downarrow$)   & \bf IS ($\uparrow$)   & \bf KL ($\downarrow$)  \\
\hline
DiffSound \cite{YangYWWWZY23}   & $50.40$ & $4.19$  & $3.63$ \\
AudioLDM-S \cite{LiuCYMLM0P23}  & $28.08$ & $6.78$ & $2.51$  \\
AudioLDM-L \cite{LiuCYMLM0P23}  & $27.51$ & $7.18$ & $2.49$ \\
AudioLDM-L-Full \cite{LiuCYMLM0P23} & 24.26 & \bf 7.67 & 2.07 \\
AudioJourney-T5 \cite{michaels2024audio} & \bf 12.09 & 1.64 & \bf 0.259 \\
\hdashline
\bf Our Spider & 27.12 & 7.36  & 2.31  \\
\hline
\end{tabular}
\caption{Text-to-Audio generation on AudioSet~\cite{gemmeke2017audio}.}
\label{tab:AudioSetResults}
\end{table}

\begin{table}[htbp]
\small
\centering
\begin{tabular}{l|cccc}
\hline
\bf Method & \bf FVD ($\downarrow$) & \bf IS ($\uparrow$) \\
\hline
CogVideo \cite{abs-2205-15868} & 702.00 & 25.27 	 \\   
MakeVideo \cite{abs-2209-14792} & 367.23 & 33.00 \\    
VideoGen \cite{li2023videogen} &  554.00 & \bf 71.61 \\
PYoCoge \cite{ge2023preserve} &  \bf 355.19 & 47.76 \\
\hdashline  
\bf Our Spider & 382.16 & 34.68 \\ 
\hline
\end{tabular}
%\vspace{1mm}
\caption{
Text-to-Video generation on UCF-101 \cite{soomro2012ucf101}.
}
\label{tab:T2V-ucf}
\end{table}

\section{Experiments of Any-to-Any Generation}
Following the setting in \cite{wu2023next}, we evaluate the performance of our Spider on various benchmark tasks, including X-to-Text generation, Text-to-X generation, and Text-conditioned modality editing.
The results in Tab.~\ref{tab:I2T-res} to Tab.~\ref{tab:TV2V-res} show that our Spider is superior to the NExT-GPT on any-to-any generation tasks, meanwhile Spider obtains competitive performance compared to the state-of-the-art methods.

\section{More Experiments on Task-specific Datasets}
\label{sec:task_spec}
Our Spider integrates the existing pre-trained models as Decoders for producing different modalities, specifically, Stable Diffusion v1.5 \cite{robin2022high} for image generation, AudioLDM \cite{LiuCYMLM0P23} for audio generation, Zeroscope v2 \cite{zeroscope} for video generation.
Thus, the modalities generation performances of our Spider are limited by the integrated Decoder models.
To improve the modalities generation performances, we can integrate more powerful Decoder models as shown in Tab.~\ref{tab:perf_comp_gen}, Tab.~\ref{tab:AudioSetResults}, and Tab.~\ref{tab:T2V-ucf}.

\noindent \textbf{Text-to-Image Generation}.
Following the setting in \cite{qu2024discriminative},
Tab.~\ref{tab:perf_comp_gen} shows the comparisons of Text-to-Image generation on COCO-NSS1K~\cite{qu2023layoutllm}, CC-500~\cite{feng2022training}, and ABC-6K~\cite{feng2022training}.
Our any-to-many Spider model obtains competitive performance compared to these state-of-the-art task-specific models.

\noindent \textbf{Text-to-Audio Generation}.
Following the setting in \cite{LiuCYMLM0P23,michaels2024audio},
Tab.~\ref{tab:AudioSetResults} shows the comparisons of Text-to-Audio generation on AudioSet~\cite{gemmeke2017audio}.
Our any-to-many Spider model obtains competitive performance compared to these state-of-the-art task-specific models.

\noindent \textbf{Text-to-Video Generation}.
Following the setting in \cite{ge2023preserve},
Tab.~\ref{tab:T2V-ucf} shows the comparisons of Text-to-Video generation on UCF-101~\cite{soomro2012ucf101}.
Our any-to-many Spider model obtains competitive performance compared to these state-of-the-art task-specific models.

\begin{table*}[!ht]
\centering
\renewcommand{\tabcolsep}{7.0pt}
\renewcommand\arraystretch{1.1}
\small
\begin{tabular}{l|ccc|ccc|ccc}
\hline
\multirow{2}*{\bf Method}  & \multicolumn{3}{c|}{\bf X-to-Text Generation} & \multicolumn{3}{c|}{\bf Text-to-X Generation} & \multicolumn{3}{c}{\bf X-to-X Generation} \\
\cline{2-10}
 & I2T ($\uparrow$) & A2T ($\uparrow$) & V2T ($\uparrow$) & T2I ($\downarrow$) & T2A ($\downarrow$) & T2V ($\downarrow$) & I2I ($\uparrow$) & A2A ($\downarrow$) & V2V ($\uparrow$)\\
\hline
NExT-GPT \cite{wu2023next} & 45.1 & 0.534 & 58.8 & 11.18 & 23.25 & 12.69 & 29.32 & 0.300 & 0.2684 \\
\hdashline
Spider (K=1) & \bf 46.1 & \bf 0.538 & \bf 60.0 & 11.18 & 23.27 & 12.66 & 29.34 & 0.299 & 0.2690 \\
Spider (K=2) & 45.9 &  0.537 & 59.8 & \bf 11.13 & 23.02 & \bf 12.62 & \bf 30.52 & 0.279 & \bf 0.2782 \\
Spider (K=3) & 45.8 & 0.534 & 59.5 & 11.13 & \bf 23.00 & 12.63 & 30.49 & \bf 0.276 & 0.2779 \\
\hline
\end{tabular}
%\vspace{-5pt}
\caption{Influence of Experts. $K$ is the number of Projection Experts in Unified Decoder Projector.}
%\vspace{-5pt}
\label{tab:ablation_expert}
\end{table*}

\begin{table*}[!ht]
\centering
\renewcommand{\tabcolsep}{7.0pt}
\renewcommand\arraystretch{1.1}
\small
\begin{tabular}{l|ccc|ccc|ccc}
\hline
\multirow{2}*{\bf Method}  & \multicolumn{3}{c|}{\bf X-to-Text Generation} & \multicolumn{3}{c|}{\bf Text-to-X Generation} & \multicolumn{3}{c}{\bf X-to-X Generation} \\
\cline{2-10}
 & I2T ($\uparrow$) & A2T ($\uparrow$) & V2T ($\uparrow$) & T2I ($\downarrow$) & T2A ($\downarrow$) & T2V ($\downarrow$) & I2I ($\uparrow$) & A2A ($\downarrow$) & V2V ($\uparrow$)\\
\hline
NExT-GPT \cite{wu2023next} & 45.1 & 0.534 & 58.8 & 11.18 & 23.25 & 12.69 & 29.32 & 0.300 & 0.2684 \\
\hdashline
Spider (w/o MRL) & 45.8 & \bf 0.538 & 59.7 & 11.14 & 23.07 & 12.62 & 29.89 & 0.289 & 0.2737\\
Spider (w MRL) & \bf 45.9 & 0.537 & \bf 59.8 & \bf 11.13 & \bf 23.02 & \bf 12.62 & \bf 30.52 & \bf 0.279 & \bf 0.2782 \\
\hline
\end{tabular}
%\vspace{-5pt}
\caption{Influence of M-Reconstruction Loss (MRL), in Decoders-Controller.}
%\vspace{-5pt}
\label{tab:ablation_reconstruct}
\end{table*}

\begin{table*}[!ht]
\centering
\renewcommand{\tabcolsep}{6.5pt}
\renewcommand\arraystretch{1.1}
\small
\begin{tabular}{l|ccc|ccc|ccc}
\hline
\multirow{2}*{\bf Method}  & \multicolumn{3}{c|}{\bf X-to-Text Generation} & \multicolumn{3}{c|}{\bf Text-to-X Generation} & \multicolumn{3}{c}{\bf X-to-X Generation} \\
\cline{2-10}
 & I2T ($\uparrow$) & A2T ($\uparrow$) & V2T ($\uparrow$) & T2I ($\downarrow$) & T2A ($\downarrow$) & T2V ($\downarrow$) & I2I ($\uparrow$) & A2A ($\downarrow$) & V2V ($\uparrow$)\\
\hline
NExT-GPT \cite{wu2023next} (Vicuna) & 45.1 & 0.534 & 58.8 & 11.18 & 23.25 & 12.69 & 29.32 & 0.300 & 0.2684 \\
\hdashline
Spider (Vicuna) & 44.8 & 0.529 & 58.9 & 11.14 & 23.04 & 12.62 & 30.49 & 0.284 & 0.2773 \\
Spider (LLaMA2) & \bf 45.9 & \bf 0.537 & \bf 59.8 & \bf 11.13 & \bf 23.02 & \bf 12.62 & \bf 30.52 & \bf 0.279 & \bf 0.2782 \\
\hline
\end{tabular}
%\vspace{-5pt}
\caption{Influence of LLM.}
%\vspace{-5pt}
\label{tab:ablation_llm}
\end{table*}

\section{More Ablation Study}
There are the notations for the ablation study: 
I2T ($\uparrow$) is Image-to-Text generation on COCO-caption \cite{LinMBHPRDZ14} with B@4 metric,
A2T ($\uparrow$) is Audio-to-Text generation on AudioCaps \cite{KimKLK19} with SPIDEr metric,
V2T ($\uparrow$) is Video-to-Text generation on MSR-VTT \cite{XuMYR16} with B@4 metric,
T2I ($\downarrow$) is Text-to-Image generation on COCO-caption \cite{LinMBHPRDZ14} with FID metric,
T2A ($\downarrow$) is Text-to-Audio generation on AudioCaps \cite{KimKLK19} with FD metric,
T2V ($\downarrow$) is Text-to-Video generation on MSR-VTT \cite{XuMYR16} with FID metric,
I2I ($\uparrow$) is Image-to-Image generation (text-conditioned image editing for object) on COCO \cite{LinMBHPRDZ14} with CLIP metric,
A2A ($\downarrow$) is Audio-to-Audio generation (text-conditioned speech editing) on VCTK \cite{veaux2017cstr} with MCD metric,
V2V ($\uparrow$) is Video-to-Video generation (text-conditioned video editing) on DAVIS \cite{PerazziPMGGS16} with CLIP-T metric.

\noindent \textbf{Influence of Experts.}
The results in Tab.~\ref{tab:ablation_expert} indicate that the number of experts $K=2$ is effective enough to allow our Unified Decoder Projector to align the LLM with multiple Decoders. 
$K=2$ obtains good performance improvements than $K=1$ on X-to-X Generation, while $K=3$ shows minor performance gains than $K=2$. Since X-to-Text Generation leverages the inherent expertise of the LLM to generate text directly, the performances of different $K$ are similar.
Since M-Prompt has a weak influence on Text-to-X task, the performances of different $K$ are similar.

\noindent \textbf{Influence of M-Reconstruction Loss.}
The results in Tab.~\ref{tab:ablation_reconstruct} show that employing MRL achieves performance improvements on X-to-X Generation, because MRL is applied to not only retain the input modality information, but also prevent ${\bar Q^X}$ from collapsing toward zero in the M-Alignment loss.
On X-to-Text Generation, the performances of Spider with and without MRL are similar, due to the inherent expertise of the LLM.
On Text-to-X Generation, the performances of Spider with and without MRL are similar, due to the T-Prompt playing a dominant role in controlling decoders.

\noindent \textbf{Influence of LLM.}
The results in Tab.~\ref{tab:ablation_llm} show that Spider with LLaMA2 achieves consistent performance improvements than using Vicuna, especially in X-to-Text Generation where the expertise of LLM plays an important role.
Note that, NExT-GPT with Vicuna achieves good performance on X-to-Text Generation, because its designed Encoder Projectors used transformer layers to produce semantic concept tokens, which can extract more detailed information of input modalities.
Our Spider only utilizes smaller linear projection layers as Encoder Projectors, and still obtains competitive results.

\begin{table*}[ht]
\centering
\fontsize{8}{13}\selectfont
\setlength{\tabcolsep}{1.5mm}
\begin{tabular}{lccccc||ccc}
\hline
\multicolumn{6}{c||}{\bf (a) Dataset Summary} & \multicolumn{3}{c}{\bf (b) Dataset Proportions} \\
\hline
\bf Dataset & \bf Data Source & \bf In$\to$Out & \bf Samples & \bf Instructions & \bf Instances & \bf Stage1 & \bf Stage2 & \bf Stage3\\ 
\hline
\multicolumn{6}{l||}{\textbf{\em $\blacktriangleright$ Existing Dataset}}\\
CC3M (I2T) & CC3M \cite{SoricutDSG18} & T+I$\to$T  & 3.3M & - & 3.3M &0.1&-&-\\
WebVid (V2T) & WebVid \cite{BainNVZ21} & T+V$\to$T  &	10M & - & 10M &0.1&-&-\\
AudioCap (A2T) & AudioCap \cite{KimKLK19} & T+A$\to$T  & 46K & - & 46K &0.1&-&-\\
\hdashline
CC3M (T2I) & CC3M \cite{SoricutDSG18} & T$\to$I  & 3.3M & - & 3.3M &0.2&-&-\\
WebVid (T2V) & WebVid \cite{BainNVZ21} & T$\to$V  &	10M & - & 10M &0.2&-&-\\
AudioCap (T2A) & AudioCap \cite{KimKLK19} & T$\to$A  & 46K & - & 46K &0.1&-&-\\
COCO (I2B) & COCO \cite{LinMBHPRDZ14} & T+I$\to$T+B  & 330K & - & 330K &0.1&-&-\\
COCO (I2M) & COCO \cite{LinMBHPRDZ14} & T+I$\to$T+M  & 330K & - & 330K &0.1&-&-\\
\hline
\multicolumn{6}{l||}{\em \textbf{$\blacktriangleright$ Our TMM Dataset}} \\
T-to-TXs (T2I) & CC3M (T2I) & T$\to$TXs  & 3.3M & 24 & 3.3M $\times$ 24 &-&0.1&0.03\\
T-to-TXs (T2V) & WebVid (T2V) & T$\to$TXs  &	10M & 24 & 10M $\times$ 24 &-&0.1&0.03\\ 
T-to-TXs (T2A) & AudioCap (T2A) & T$\to$TXs  & 46K & 24 & 46K $\times$ 24 &-&0.1&0.03\\
\hdashline
X-to-TXs (I2T) & CC3M (I2T) & T+I$\to$TXs  & 3.3M & 17 & 3.3M $\times$ 17 &-&0.2&0.06\\
X-to-TXs (V2T) & WebVid (V2T) & T+V$\to$TXs  &	10M & 17 & 10M $\times$ 17 &-&0.2&0.06\\
X-to-TXs (A2T) & AudioCap (A2T) & T+A$\to$TXs  & 46K & 17 & 46K $\times$ 17 &-&0.1&0.03\\
X-to-TXs (I2B) & COCO (I2B) & T+I$\to$TXs  & 330K & 18 & 330K $\times$ 18 &-&0.1&0.03\\
X-to-TXs (I2M) & COCO (I2M) & T+I$\to$TXs  & 330K & 16 & 330K $\times$ 16 &-&0.1&0.03\\
\hdashline
T-to-TXs SmMI (T2V) & WebVid (T2V) & T$\to$TXs  &	10M & 24 & 10M $\times$ 24 &-&-&0.5\\ 
T-to-TXs SpMI (T2V) & WebVid (T2V) & T$\to$TXs  &	10M & 5 & 10M  $\times$ 5 &-&-&0.1\\ 
T-to-TXs TGI (GPT-4o) & GPT-4o & T$\to$TXs  & 1000 & 6 & 1000  $\times$ 6 &-&-&0.1\\ 
\hline
\multicolumn{6}{l||}{\em \textbf{$\blacktriangleright$ Our Pseudo X-to-Xs Dataset}}\\
Pseudo T-to-Xs (T2I) & T-to-TXs (T2I) & T$\to$Xs  & 2,000 & - & 2,000 &-&-&-\\
Pseudo T-to-Xs (T2V) & T-to-TXs (T2V) & T$\to$Xs  &	2,000 & - & 2,000 &-&-&-\\
Pseudo T-to-Xs (T2A) & T-to-TXs (T2A) & T$\to$Xs  & 2,000 & - & 2,000 &-&-&-\\
\hdashline
Pseudo X-to-Xs (I2T) & X-to-TXs (I2T) & T+I$\to$Xs  & 2,000 & - & 2,000 &-&-&-\\
Pseudo X-to-Xs (V2T) & X-to-TXs (V2T) & T+V$\to$Xs  &	2,000 & - & 2,000 &-&-&-\\
Pseudo X-to-Xs (A2T) & X-to-TXs (A2T) & T+A$\to$Xs  & 2,000 & - & 2,000 &-&-&-\\
Pseudo X-to-Xs (I2B) & X-to-TXs (I2B) & T+I$\to$Xs  & 2,000 & - & 2,000 &-&-&-\\
Pseudo X-to-Xs (I2M) & X-to-TXs (I2M) & T+I$\to$Xs  & 2,000 & - & 2,000 &-&-&-\\
\hdashline
Pseudo T-to-Xs SmMI (T2V) & T-to-TXs SmMI (T2V) & T$\to$Xs  &	2,000 & - & 2,000 &-&-&-\\
Pseudo T-to-Xs SpMI (T2V) & T-to-TXs SpMI (T2V) & T$\to$Xs  &	2,000 & - & 2,000 &-&-&-\\
Pseudo T-to-Xs TGI (GPT-4o) & T-to-TXs TGI (GPT-4o) & T$\to$Xs  & 1000 & - & 1000 &-&-&-\\
\hline
\end{tabular}
% \vspace{-1mm}
\caption{
(a) Dataset summary.
In$\to$Out denotes input to output modality.
Samples are the amount of unique In$\to$Out modality pair, e.g., CC3M contains 3.3M T+I$\to$T modality pairs, and T-to-TXs (T2I) contains 3.3M T$\to$TXs modality pairs constructed from the T2I data source CC3M.
Instructions are the amount of constructed user instructions, where each modality pair data can be used to build many corresponding user instructions.
Instances are the maximum amount of pairs of user instruction and answer, which equals to the amount of Samples $\times$ Instructions.
T-to-TXs SmMI (T2V) denotes T-to-TXs Smart-Multimodal Instruction (WebVid).
T-to-TXs SpMI (T2V) denotes T-to-TXs Specific-Multimodal Instruction (WebVid).
T-to-TXs TGI (GPT-4o) denotes T-to-TXs Travel-Guide Instruction (GPT-4o).
T: Text, I: Image, V: Video, A: Audio,
B: Bounding Box, M: Mask.
(b) Dataset proportions in different stages of Spider training. Stage1: X-to-X Pretraining, Stage2: X-to-TXs Finetuning, Stage3: X-to-TXs Instruction Finetuning.
}
\vspace{-3mm}
\label{tab:dataset}
\end{table*}

\section{Existing Dataset}
\noindent \textbf{CC3M \cite{SoricutDSG18} (Image-Text) Dataset} is a large-scale collection of approximately 3.3 million image-text pairs, curated from web sources. It features automatically generated captions that describe diverse visual content, making it ideal for tasks like image captioning, text-to-image retrieval, and multimodal pretraining. Despite some noise in the data due to automated generation, CC3M's scale and diversity make it a foundational resource for multimodal AI research.

\noindent \textbf{Webvid \cite{BainNVZ21} (Video-Text) Dataset} is a large-scale collection of video-text pairs designed for training and evaluating video-language models. It contains over 10 million video-text pairs sourced from the web, covering diverse topics, scenes, and activities. The dataset includes short videos with automatically generated captions, providing a rich resource for tasks like video captioning, text-to-video retrieval, and video-language pretraining. Its extensive scale and diversity make WebVid a critical resource for advancing research in video-language understanding and generation.

\noindent \textbf{AudioCap \cite{KimKLK19} (Audio-Text) Dataset} is a comprehensive dataset designed for audio captioning, enabling AI models to learn and generate natural language descriptions for diverse audio events. Built upon the AudioSet dataset, it contains over 46,000 YouTube audio clips, each paired with human-annotated captions. With a broad range of sounds, from environmental noises to musical and human activities, AudioCaps serves as a benchmark for audio-text understanding. Additionally, it features over 5,000 clips with multiple captions, supporting research into linguistic diversity in audio descriptions and multimodal learning.

\noindent \textbf{COCO \cite{LinMBHPRDZ14} (Image-Box and Image-Mask) Dataset} is a widely-used dataset for computer vision tasks, providing over 330,000 images annotated with detailed object bounding boxes, segmentation masks and captions. Designed to support object detection, segmentation, and captioning, COCO includes 80 object categories with diverse real-world scenes containing multiple objects in context. Its high-quality annotations and diverse visual content make it a fundamental benchmark for advancing multimodal AI research and training vision-language models.

\section{TMM Dataset}
\label{sec:tmm_dataset}
We constructed a new Text-formatted Many-Modal (TMM) dataset to train the Spider model, enabling it to learn the X-to-Xs capability, i.e., to achieve Any-to-Many Modalities Generation. 
In the TMM dataset, the input is in the form of 'Text' or 'Text + X', which follows the Input Question Format. The output is in the form of \textit{Text-formatted Xs (TXs)}, that is text, containing many-modal signal prompts.
As illustrated in Fig.~\ref{fig:template} (a), the Output Answer Format is the TXs format.
Eventually, the TMM dataset contains three types of datasets for different usage in training: \textit{T-to-TXs} dataset for T-to-Xs capability finetuning, \textit{X-to-TXs} dataset for X-to-Xs capability finetuning, and \textit{T-to-TXs instruction} dataset for T-to-Xs instruction finetuning.

We show some examples in Fig.~\ref{fig:tmm}.
The construction details will be made public in our source code. 
The statistic details are shown in Tab.~\ref{tab:dataset} (a).
Note that, Instances are the maximum amount of pairs of user instruction and answer, which equals to the amount of Samples $\times$ Instructions.
In the training process, each Instance is built online by combining a selected Sample and one of the Instructions.
Specifically, for each dataset construction process: (a) Pre-define an instruction pool that contains diverse Instructions referring to specific modality combinations such as image, video, audio, box, and mask. An example of the instruction template is "Please generate an image and a video based on the following text: \{\}", where \{\} is the placeholder for the text content.
(b) Randomly select an instruction template from the instruction pool to construct the question. An example of the question is "[INPUT] [SMARTMULTIMODAL] Please generate an image and a video based on the following text: A cat is sitting on a couch".
(c) Parse the instruction template to identify target modalities (i.e., image and video for the given example), then construct the answer containing the target modalities. An example of the answer is "[OUT] A cat is sitting on a couch. $<$IMAGE$>$ A cat is sitting on a couch [IMAGE$_0$] $<$/IMAGE$>$. $<$VIDEO$>$ A cat is sitting on a couch [VIDEO$_0$] $<$/VIDEO$>$ [END]".

\noindent \textbf{T-to-TXs Dataset} contains three sub-datasets constructed from CC3M \cite{SoricutDSG18} (Image-Text), AudioCap \cite{KimKLK19} (Audio-Text), and Webvid \cite{BainNVZ21} (Video-Text).
We design corresponding Any-to-Many Instruction Templates for each task to construct the T-to-TXs dataset, resulting in T-to-TXs (T2I), T-to-TXs (T2V), and T-to-TXs (T2A).

\noindent \textbf{X-to-TXs Dataset} consists of five sub-datasets constructed from CC3M \cite{SoricutDSG18} (Image-Text), COCO \cite{LinMBHPRDZ14} (Image-Box and Image-Mask), AudioCap \cite{KimKLK19} (Audio-Text), and Webvid \cite{BainNVZ21} (Video-Text).
We design corresponding Any-to-Many Instruction Templates for each task to construct the X-to-TXs dataset, resulting in X-to-TXs (I2T), X-to-TXs (V2T), X-to-TXs (A2T), X-to-TXs (I2B), and X-to-TXs (I2M).

\noindent \textbf{T-to-TXs Instruction Dataset} contains three sub-datasets which are constructed following the Any-to-Many Instruction Template format, including the smart-multimodal sub-dataset named T-to-TXs SmMI (T2V), the specific-multimodal instruction sub-dataset named T-to-TXs SpMI (T2V), and the travel-guide instruction sub-dataset named T-to-TXs TGI (GPT-4o).
The T-to-TXs SmMI (T2V) and T-to-TXs SpMI (T2V), concatenate multiple samples from Webvid, to mimic the arbitrary combination of output modalities.
The T-to-TXs TGI (GPT-4o) is constructed with the assistance of GPT-4o, including 1000 travel guides for cities around the world.

\section{Pseudo X-to-Xs Dataset}
\label{sec:pseudo_dataset}
We use the Spider model well-trained on the TMM (X-to-TXs) dataset to generate a new pseudo X-to-Xs dataset. 
This is a first-ever X-to-Xs many-modal dataset for the Any-to-Many Modalities Generation task, providing rich data support for future research.
The output form of TMM (X-to-TXs) dataset is TXs (i.e., text only) without diverse modalities, while the pseudo X-to-Xs dataset contains arbitrary combination modalities.
With TMM (X-to-TXs) dataset, our Spider is able to perform X-to-Xs generation, due to no need to train the multimodal Decoders.
With the pseudo X-to-Xs dataset, the multimodal Decoders can be end-to-end fine-tuning with LLM if needed in future work, due to having the ground truth modalities to supervise the Decoders.
The statistic details are shown in Tab.~\ref{tab:dataset} (a).

\begin{figure*}[t]
%\vspace{-2mm}
\centering
\includegraphics[width=1.0\textwidth]{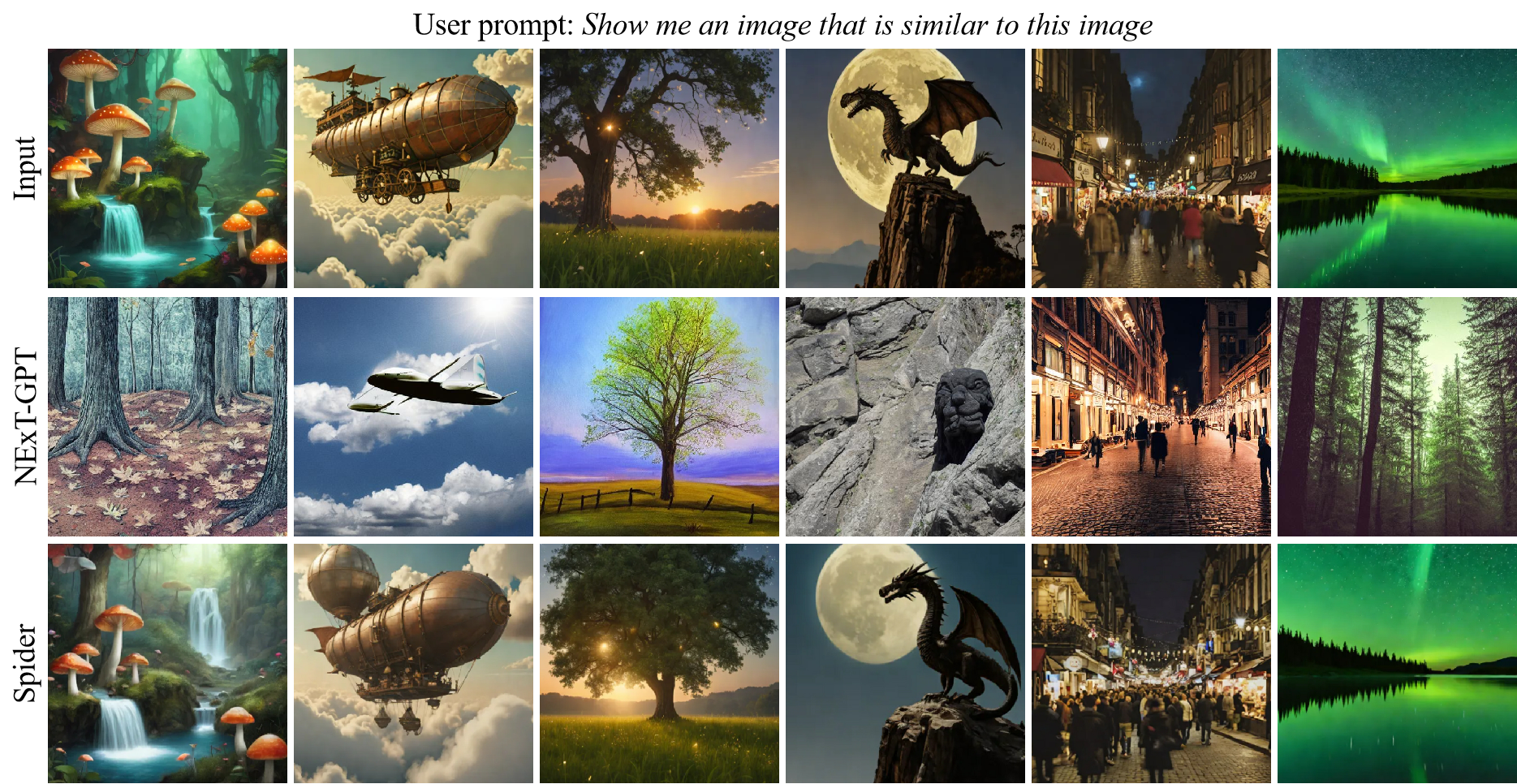}
\caption{\textbf{Text + Image $\to$ Text + Image}.
User prompt is "Show me an image that is similar to this image". Spider generated the image that is more similar to the input image, compared to NExT-GPT.
}
%\vspace{-2mm}
\label{fig:example1}
\end{figure*}

\section{Spider Training}
\label{sec:spider_train}
The training process of our Spider model consists of three stages, including X-to-X Pretraining (Stage1), X-to-TXs Finetuning (Stage2), and X-to-TXs Instruction Finetuning (Stage3).
The dataset proportions in different training stages are shown in Tab.~\ref{tab:dataset} (b).

\noindent \textbf{X-to-X Pretraining} enables Spider to perform the basic X-to-X generation, connecting the four parts of Spider including Encoders, LLM, Decoders-Controller, and Decoders. 
As shown in Fig.\ref{fig:spider}, we only train the input-side Encoder Projectors, LoRA of LLM, and output-side Decoders-Controller, while other parameters are frozen.
In this stage, we employ X-to-X tasks for training, including X-to-text generation, text-to-X generation, Image-to-Box prediction, and Image-to-Mask prediction:
(a) X-to-text generation encompasses tasks such as image, video, and audio captioning, where the model is trained to produce textual descriptions from multimodal inputs. 
In this process, the input-side Encoder Projectors are trained to align the output embedding of modality-specific Encoders and the textual embedding space of pre-trained LLM. It enables the LLM to understand the input modalities encoded by the Encoders.
(b) Text-to-X generation (i.e., video, image, and audio generation), Image-to-Box prediction, and Image-to-Mask prediction, aim at aligning the output textual embedding space of LLM and the input end of modality-specific Decoders, where the output-side Decoders-Controller is trained via the M-Alignment Loss and M-Reconstruction Loss. The output embedding of LLM is projected by the Decoders-Controller into the controlling embedding, which can control the Decoders for modalities generation.

\noindent \textbf{X-to-TXs Finetuning} enables Spider to have the basic ability of X-to-Xs generation, via finetuning the LoRA of LLM with the proposed T-to-TXs and X-to-TXs Datasets.
After the Stage1 X-to-X Pretraining, Spider is able to perform the basic X-to-X generation, but the LLM can only produce the single-modal signal prompt to control X-modality generation.
In order to perform X-to-Xs generation, we finetune the LoRA of LLM with the proposed T-to-TXs and X-to-TXs Datasets, allowing the LLM to produce many-modal signal prompts to control Xs-modalities generation.
In this stage, the model is trained to produce TXs containing many-modal signal prompts, i.e., pure textual outputs.

\noindent \textbf{X-to-TXs Instruction Finetuning} makes Spider achieve X-to-Xs generation in a proper manner, i.e., faithfully following and understanding the user instructions and generating desired many-modal outputs. We further finetune the LoRA of LLM using the proposed T-to-TXs Instruction Dataset, T-to-TXs and X-to-TXs Datasets.
For each iteration, 70\% data are from T-to-TXs Instruction Dataset, and 30\% are from T-to-TXs and X-to-TXs Datasets.

\begin{table}[!th]
\centering
\fontsize{8}{11}\selectfont
\setlength{\tabcolsep}{3.5mm}
\begin{tabular}{lccc}
\hline
\bf Configuration & \bf Stage1 & \bf Stage2 & \bf Stage3 \\ 
\toprule
Optimizer & Adam & Adam & Adam  \\
Learning Rate & 0.0001 & 0.0001 & 0.0001\\
Weight Decay & 0.001 & 0.001 & 0.001\\
Iterations Per Epoch & 10k & 1k & 1k\\
Training Epochs & 40 & 20 & 20\\
Batch Size Per GPU & 4 & 4 & 4\\
Max Token Length & 1024 & 1024 & 1024\\
Freeze LLM & No  & No & No \\
\hline
\end{tabular}
\caption{
Training configurations of Spider.
}
\vspace{-1mm}
\label{tab:training_recipes}
\end{table}

\section{Training Configurations}
Table \ref{tab:training_recipes} shows the training configurations of Spider.
%Besides, we freeze the pre-trained text-input projector and LLM output-head, to alleviate the catastrophic forgetting problem of the pre-trained LLM. It is helpful to retain the text ability of the pre-trained LLM, accelerating the model convergence.

\section{Qualitative Analysis}
\label{sce:vis}
\begin{figure}[ht]
%\vspace{-2mm}
\centering
\includegraphics[width=1.0\columnwidth]{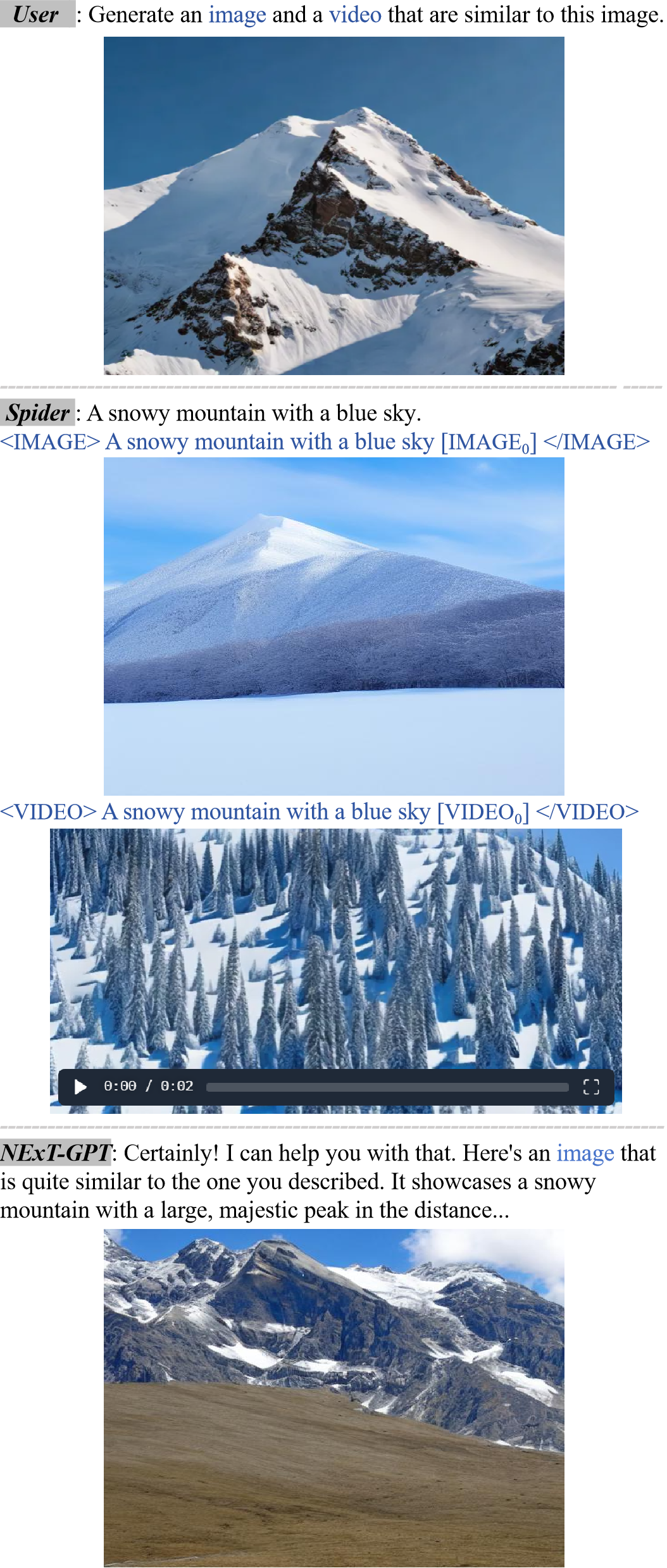}
\caption{\textbf{Text + Image $\to$ Text + Image + Video}.
User prompt is "Generate an image and a video that are similar to this image."
Spider generated an image and a video according to the user prompt. But NExT-GPT only generated an image, and failed to generate the video.
}
%\vspace{-2mm}
\label{fig:example5}
\end{figure}

\begin{figure}[ht]
%\vspace{-2mm}
\centering
\includegraphics[width=1.0\columnwidth]{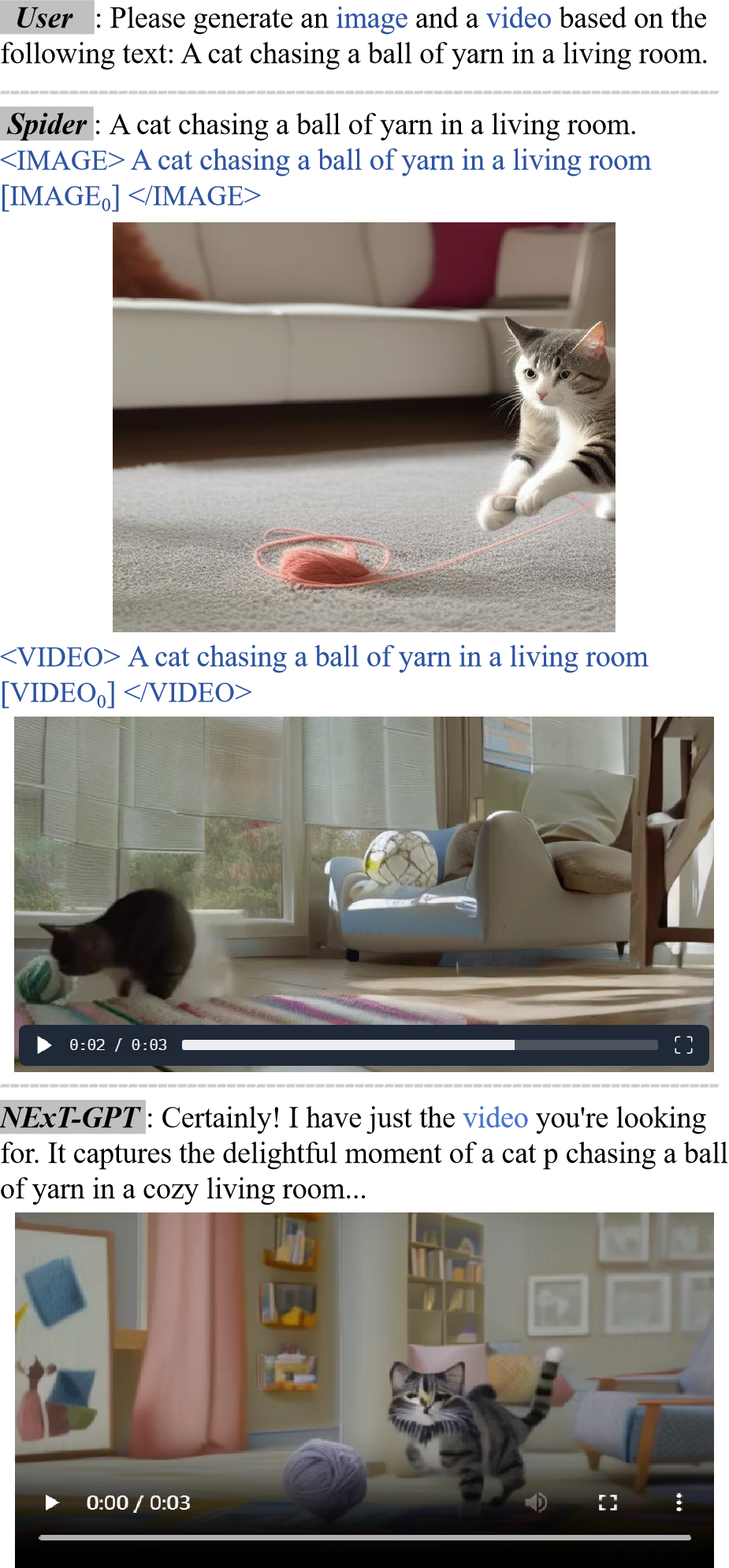}
\caption{\textbf{Text $\to$ Text + Image + Video}.
User prompt is "Please generate an image and a video based on the following text: A cat chasing a ball of yarn in a living room."
Spider generated an image and a video according to the user prompt. But NExT-GPT only generated a video, and failed to generate the image.
}
%\vspace{-2mm}
\label{fig:example4}
\end{figure}

\begin{figure}[ht]
%\vspace{-2mm}
\centering
\includegraphics[width=1.0\columnwidth]{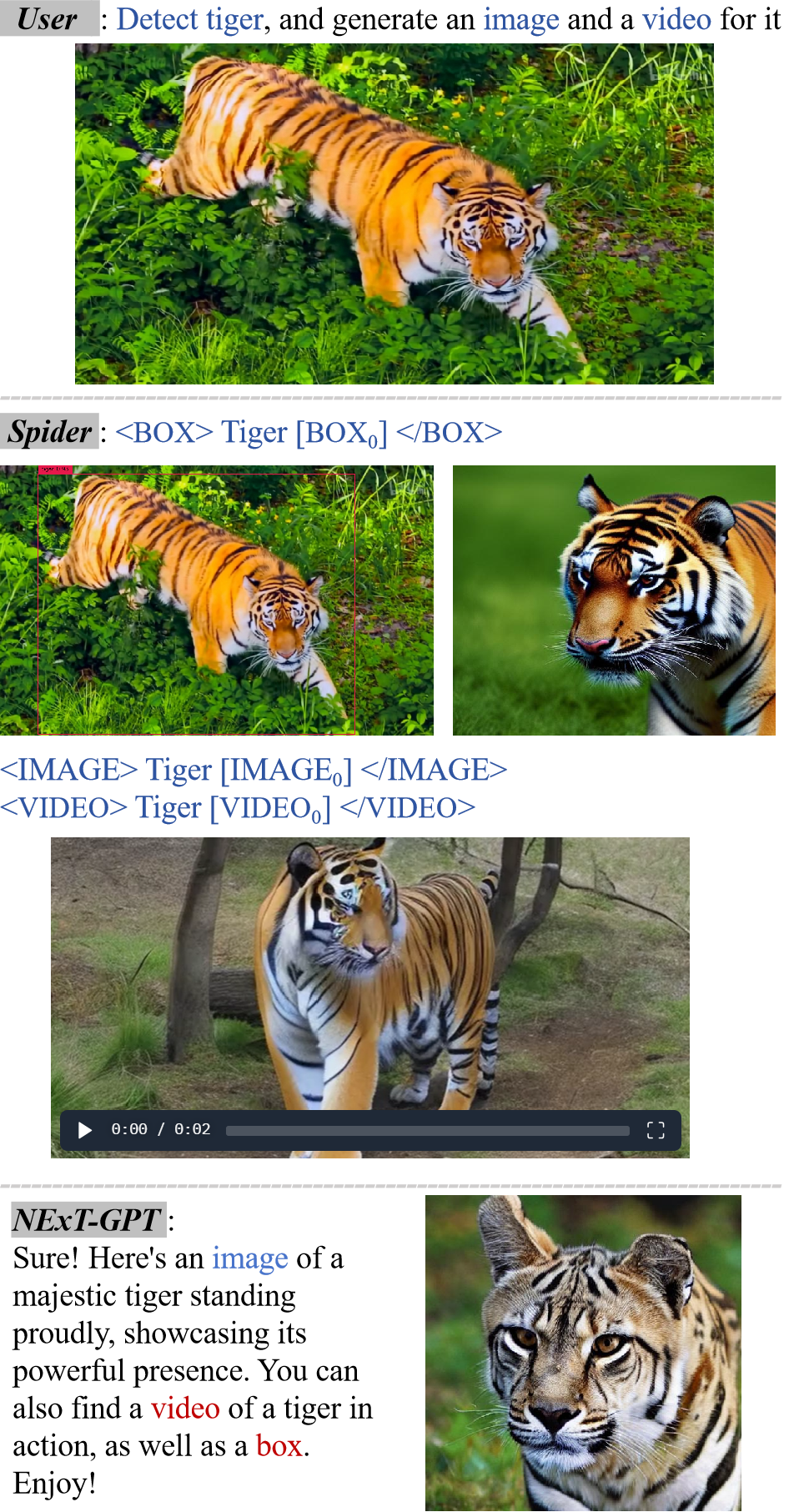}
\caption{\textbf{Text + Image $\to$ Box + Image + Video}.
User prompt is "Detect tiger, and generate an image and a video for it". Spider detected the tiger in the input image, and generated an image and a video according to the user prompt. But NExT-GPT only generated an image, failed to generate the video and detect the tiger in the input image. NExT-GPT mentioned the video and box in the text response, but it still failed to generate them.
}
%\vspace{-2mm}
\label{fig:example6}
\end{figure}

\begin{figure}[ht]
%\vspace{-2mm}
\centering
\includegraphics[width=1.0\columnwidth]{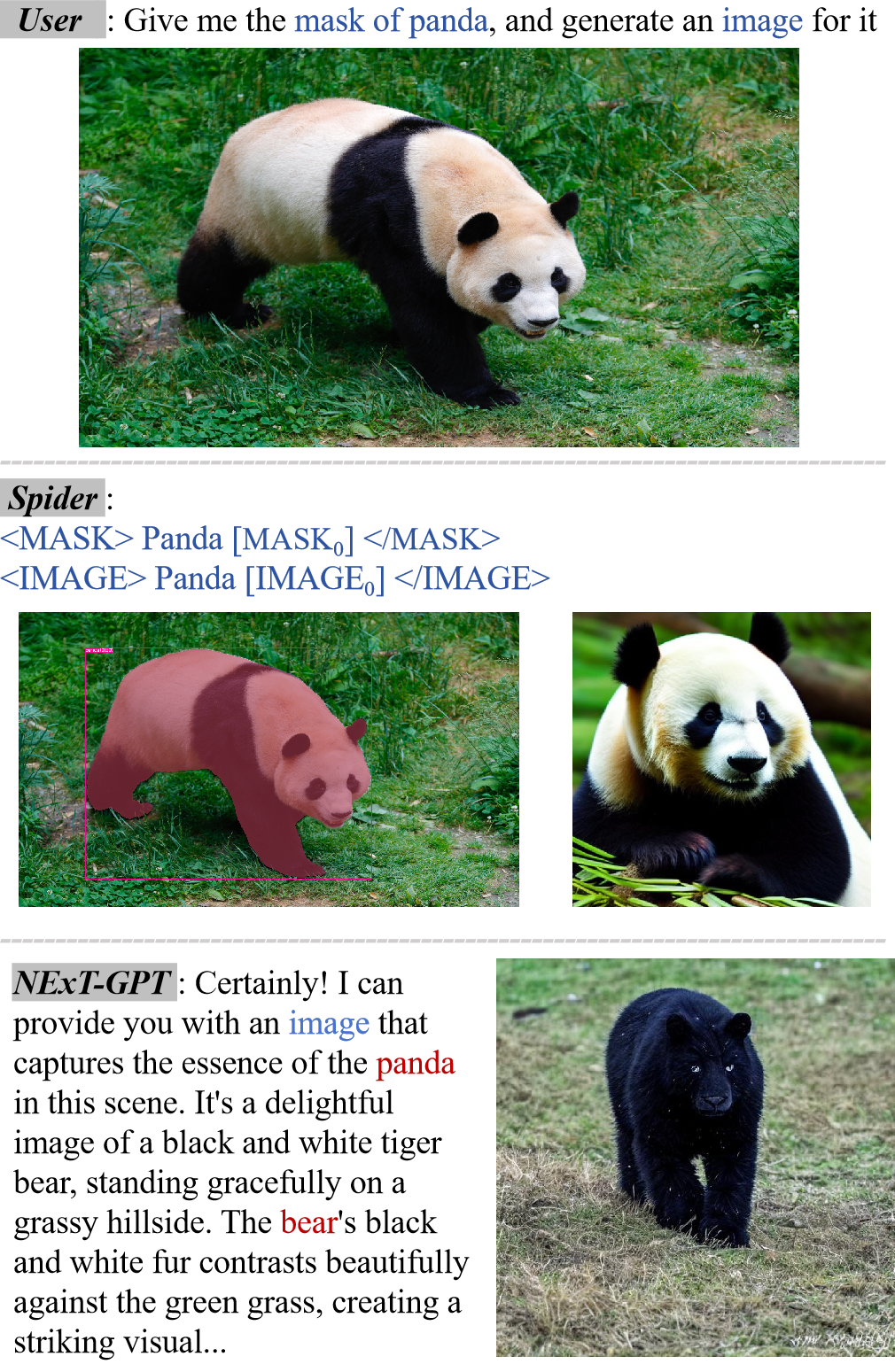}
\caption{\textbf{Text + Image $\to$ Mask + Image}.
User prompt is "Give me the mask of panda, and generate an image for it". Spider generated the mask for the panda in the input image, and generated an image according to the user prompt. But NExT-GPT wrongly generated an image of bear, and failed to generate the mask for the panda in the input image.
}
%\vspace{-2mm}
\label{fig:example7}
\end{figure}

\begin{figure}[ht]
%\vspace{-2mm}
\centering
\includegraphics[width=1.0\columnwidth]{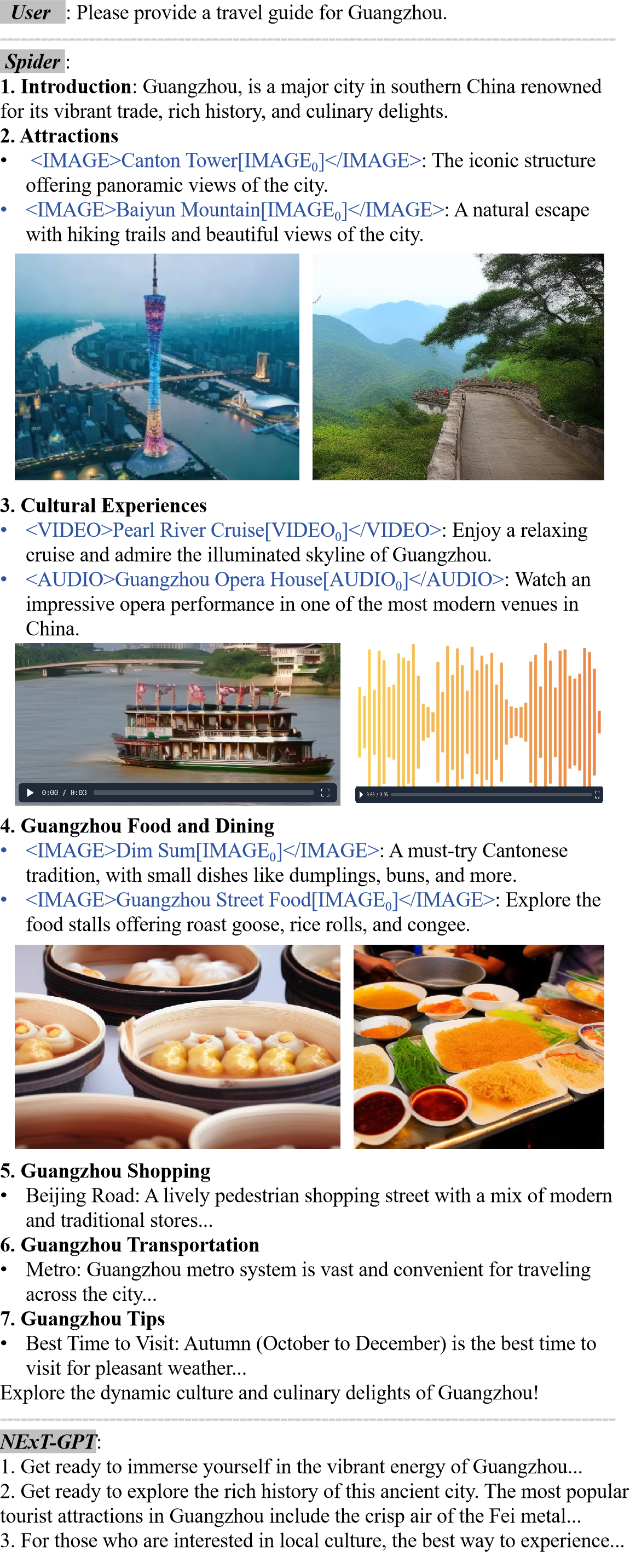}
\caption{\textbf{Text $\to$ Text + Image + Video + Audio}. Spider generated a many-modal travel guide. But NExT-GPT only generated a textual travel guide.}
%\vspace{-2mm}
\label{fig:example2}
\end{figure}

\begin{figure}[ht]
%\vspace{-2mm}
\centering
\includegraphics[width=1.0\columnwidth]{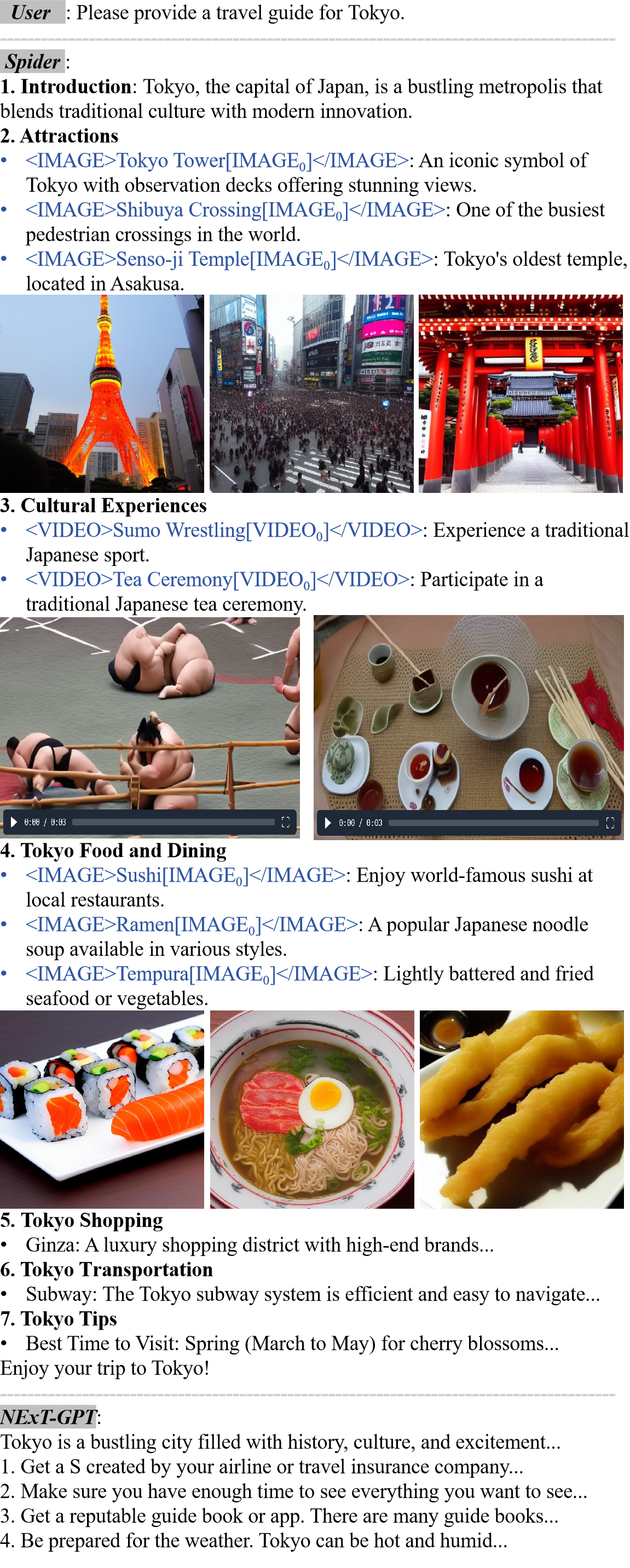}
%\vspace{-4mm}
\caption{\textbf{Text $\to$ Text + Image + Video + Audio}. Spider generated a many-modal travel guide. But NExT-GPT only generated a textual travel guide.}
%\vspace{-2mm}
\label{fig:example3}
\end{figure}

We provide comparisons in Fig.~\ref{fig:example1}, \ref{fig:example5}, \ref{fig:example4}, \ref{fig:example6}, \ref{fig:example7}, \ref{fig:example2}, \ref{fig:example3}, to demonstrate Spider's remarkable ability to generate arbitrary combinations of modalities within a single response.

%\iffalse

%\fi

\end{document}